\documentclass[11pt]{article}

\usepackage[]{ACL2023}

\usepackage{times}
\usepackage{latexsym}
\usepackage{longtable}

\usepackage[T1]{fontenc}
\usepackage[utf8]{inputenc}

\usepackage{microtype}

\usepackage{inconsolata}

\usepackage{microtype}
\usepackage{graphics}
\usepackage{graphicx}
\usepackage{turnstile}

\newcommand{\ourdataset}{Synthetic-Persona-Chat}
\newcommand{\personachat}{Persona-Chat}
\newcommand{\ourdatasetabbr}{SPC}
\newcommand{\personachatabbr}{PC}
\newcommand{\ourdatasetlink}{https://github.com/google-research-datasets/Synthetic-Persona-Chat}
\newcommand{\commentout}[1]{%
}
\usepackage{multirow}

\usepackage{amsmath}

\usepackage{enumitem}
\usepackage{array}
\usepackage{booktabs}
\usepackage{amssymb}
\newcolumntype{C}[1]{>{\centering\arraybackslash}p{#1}}

\title{Faithful Persona-based Conversational Dataset Generation with Large Language Models}

\author{Pegah Jandaghi \thanks{~~ Work done during an internship at Google Inc., Mountain View, USA}\\
  University of Southern California \\
  \texttt{jandaghi@usc.edu} \\\And
  XiangHai Sheng \\
  Google  \\
  \texttt{xhs@google.com} \\ \And
  Xinyi Bai \\
  Google \\
  \texttt{shinii@google.com} \\ \AND
  Jay Pujara \\
  Information Sciences Institute  \\
  \texttt{jpujara@isi.edu} \\ \And
  Hakim Sidahmed \\
  Google Research \\
  \texttt{hsidahmed@google.com}
  }
\begin{document}
\maketitle
\begin{abstract}
High-quality conversational datasets are essential for developing AI models that can communicate with users.
One way to foster deeper interactions between a chatbot and its user is through \emph{personas}, aspects of the user's character that provide insights into their personality, motivations, and behaviors.
Training Natural Language Processing (NLP) models on a diverse and comprehensive persona-based dataset can lead to conversational models that create a deeper connection with the user, and maintain their engagement. 
In this paper, we leverage the power of Large Language Models (LLMs) to create a large, high-quality conversational dataset from a seed dataset. 
We propose a Generator-Critic architecture framework to expand the initial dataset, while improving the quality of its conversations.
The Generator is an LLM prompted to output conversations.
The Critic consists of a mixture of expert LLMs that control the quality of the generated conversations.
These experts select the best generated conversations, which we then use to improve the Generator.
We release \ourdataset{}\footnote{Dataset available at \ourdatasetlink}, consisting of 20k conversations seeded from \personachat{} \cite{Zhang2018PersonalizingDA}.
We evaluate the quality of \ourdataset{} and our generation framework on different dimensions through extensive experiments, and observe that the losing rate of \ourdataset{} against \personachat{} during Turing test decreases from $17.2\%$ to $8.8\%$ over three iterations.
\end{abstract}
\section{Introduction}
% jay comments: write your goals, justification for goals, my choices for method, justification for my method

% why personas matter?"
% personas are everywhere and how we interrelated with people, ... marketing, system design, health --> This general
\commentout{
As technology continues to shape our lives and influence the way we interact with products and services, the need to understand users on a deeper level becomes increasingly crucial. One effective approach to gaining insight into user behaviour is the use of personas. Personas, which are abstract user representations, have become prevalent in a variety of domains including marketing, system design, and healthcare \cite{Pruitt2003PersonasPA}. Previous persona based conversations have low quality which is reflected in the models trained on them and are not scaleable. In particular, these datasets lack \emph{faithfulness} to personas. Moreover, the constant evolution of personas and models requires a living dataset. 
As a result, there is a need for more advanced methods of generating persona-based conversational datasets that can overcome these limitations.
In this paper, we address these limitation via a system that uses LLMs and creates high quality datasets. 
}

Every person is a story. 
Systems that interact with people must understand their underlying stories to effectively engage with them.
Unfortunately, many existing datasets used for training conversational agents do not sufficiently model their users.
\emph{Personas} - abstract user representations that express the ``story'' of a person based on their background and preferences - have been widely used for human-centered design in a variety of domains, including marketing, system design, and healthcare \cite{Pruitt2003PersonasPA}.
Prior persona-based conversational datasets, like \personachat~(\personachatabbr{}) \cite{Zhang2018PersonalizingDA}, suffer from several limitations, such as small size, static dialogues that cannot easily be updated with new topics, irrelevant utterances, and contradictory \emph{persona attributes} \cite{Wu2019GettingTK}.
In this paper, we propose a novel framework for generating large, dynamic, persona-based conversational datasets that capture the breadth and depth of human experience.

% What are use cases?
% use cases:
% communicate with workers
% communicate and create a narrative for patient/educational messages
% better outreach for market
%\pegah{what are personas, where are they used and how}
Personas~\cite{10.1145/997078.997089, 10.5555/553473} have been widely used in a variety of domains and applications, including creating narratives for patients and sharing educational messages in healthcare~\cite{Massey2021vaccine}, targeting users in marketing~\cite{Pinxteren2020, Fuglerud2020}, and communicating with workers in management \cite{Claus2019Management}. Conversational agents use personas to generate more interesting and engaging conversations with their users \cite{zhou2019design, Shum2019SketchFillARAP}.
% \cite{} introduced Persona-chat which consists of persona-based conversations created by humans. This dataset has been widely used however it has low quality and the generation by humans is not scalable.
% \pegah{mention the types of quality issues. maybe describe more on specificity} 

Creating persona-based datasets is difficult: the process is labor-intensive, the outputs must be updated to reflect current events and new concepts, and there are often quality concerns.
Existing persona-based datasets have resulted from labor-intensive data collection processes \cite{Zhang2018PersonalizingDA, Zhong2020EndowingED} involving 
% Existing persona-based datasets have been the result of a labor-intensive data collection process \cite{Zhang2018PersonalizingDA, Zhong2020EndowingED} that have involved 
humans to create or validate personas, create fictional persona-based conversations, and ensure the conversations are coherent.
Moreover, even after these datasets are created, it is difficult to update them with the latest topics~\cite{lee-etal-2022-personachatgen}, such as current events, new concepts, products, or social trends \cite{Lazaridou2021MindTG}. Finally, existing persona-based datasets do not guarantee \emph{faithfulness}, a criterion we introduce to describe the alignment between participants' utterances and their personas.
%Finally, and to the best of our knowledge, none of the existing persona-based conversational datasets address the importance to generate faithful conversations.
%Faithfulness, formally defined in the next sections, indicates how much the downstream tasks can trust the inferred personas from a conversation. 

In this paper, we introduce a new framework for generating large, customized persona-based conversational datasets that uses unsupervised LLMs to reduce human labor, introduces methods to generate, expand, and update personas automatically, and enforces a set of quality criteria including faithfulness to ensure dialogues are human-like.
% \cite{lee-etal-2022-personachatgen} addressed some of persona-Chat limitations by suggesting to generate similar dataset using LMs. 
% However this dataset relies on human experts for generating personas and its quality is as good as the quality of the LM. 
%We propose an unsupervised method to generate user personas, combined with a few-shot conversation generation framework to address the above mentioned issues.
Our persona-based conversational dataset generation framework consists of a three-level pipeline:
\begin{enumerate}
    \item User Generation
    \item User Pairing
    \item Conversation Generation
\end{enumerate}
The user generation step takes a set of seed personas, and augments it to create plausible user profiles.
The user pairing step matches users to participate in conversations.
The conversation generation produces plausible conversations between the selected user pairs.
The conversation generation component uses a method similar to self-feedback  \cite{madaan2023selfrefine} to iteratively improve the quality of generated samples.
% limitations:
% diversity
% specificity
% faithfulness
% time(deprecation)
% Experimental psychology studies suggest that if we know enough about an individual’s characteristics and goals, then we can predict their responses to different scenarios.
% Therefore they are used as a medium for communication.
%  Which will lead to better narrative in conversation.
% Why personas should be deep?
% Research shows that generalisations and stereotypes in personas may only be effective for short-term communication, and richer personas should be employed for long-term and more engaging use. \cite{inbook}
% % Why personas should be diverse? 
% \cite{10.1145/3290605.3300565}
% \pegah{A paragraph on going from personas to persona based conversations}
% Jay: There is a need for such dataset should be justified. 
% What are the issues with previous datasets?:
% generic personas --> not covering different features..
% contradicting personas 
% fluency? & naturalness?
% inference issues?
% scalability & flexibility

We used the proposed framework to create \ourdataset{} (\ourdatasetabbr{}), a conversational dataset with $5k$ user personas, and $20k$ faithful dialogues.
The framework we defined to create this dataset can be reused to define specialized personas, such as user music profiles, etc. to create application-specific datasets.

Our contributions are:
\begin{itemize}
% [leftmargin=*]
\item We propose an unsupervised approach to generate, and extend specialized personas using LLMs.
\item We introduce and evaluate a framework based on LLMs to evolve a dataset while imposing different objectives on it. 
\item We release \ourdataset{}, a high-quality, faithful, persona-based conversational dataset useful for several conversational tasks, such as training persona inference models.
\end{itemize}

% These instructions are for authors submitting papers to ACL 2023 using \LaTeX. They are not self-contained. All authors must follow the general instructions for *ACL proceedings,\footnote{\url{http://acl-org.github.io/ACLPUB/formatting.html}} as well as guidelines set forth in the ACL 2023 call for papers.\footnote{\url{https://2023.aclweb.org/calls/main_conference/}} This document contains additional instructions for the \LaTeX{} style files.
% The templates include the \LaTeX{} source of this document (\texttt{acl2023.tex}),
% the \LaTeX{} style file used to format it (\texttt{acl2023.sty}),
% an ACL bibliography style (\texttt{acl\_natbib.bst}),
% an example bibliography (\texttt{custom.bib}),
% and the bibliography for the ACL Anthology (\texttt{anthology.bib}).

% \section{Engines}

% To produce a PDF file, pdf\LaTeX{} is strongly recommended (over original \LaTeX{} plus dvips+ps2pdf or dvipdf). Xe\LaTeX{} also produces PDF files, and is especially suitable for text in non-Latin scripts.
% \begin{table}
% \centering
% \begin{tabular}{lc}
% \hline
% \textbf{Command} & \textbf{Output}\\
% \hline
% \verb|{\"a}| & {\"a} \\
% \verb|{\^e}| & {\^e} \\
% \verb|{\`i}| & {\`i} \\ 
% \verb|{\.I}| & {\.I} \\ 
% \verb|{\o}| & {\o} \\
% \verb|{\'u}| & {\'u}  \\ 
% \verb|{\aa}| & {\aa}  \\\hline
% \end{tabular}
% \begin{tabular}{lc}
% \hline
% \textbf{Command} & \textbf{Output}\\
% \hline
% \verb|{\c c}| & {\c c} \\ 
% \verb|{\u g}| & {\u g} \\ 
% \verb|{\l}| & {\l} \\ 
% \verb|{\~n}| & {\~n} \\ 
% \verb|{\H o}| & {\H o} \\ 
% \verb|{\v r}| & {\v r} \\ 
% \verb|{\ss}| & {\ss} \\
% \hline
% \end{tabular}
% \caption{Example commands for accented characters, to be used in, \emph{e.g.}, Bib\TeX{} entries.}
% \label{tab:accents}
% \end{table}
\section{Definitions}
We define the faithful persona-based dialogue generation task. We begin by defining the persona-based dialogue generation task.
We then formally define the faithfulness criteria as a desired quality for the generated dialogues.
Throughout this section, we use $\pi$ to refer to persona attributes (individual sentences which, together, form the user persona), $U$ to refer to user profiles, and $D$ to refer to conversations (dialogues).

\textbf{Persona Attributes}
% We follow \citet{Zhang2018PersonalizingDA} in defining a user
We define a user persona attribute as a sentence describing this user.
"I like ice cream", "I have two brothers" and "My native language is Tamazight" are all examples of persona attributes.
Let $\Omega$ be the universal set of persona attributes.
$\Omega$ contains all natural language descriptions of all tangible features of any person, which is unbounded.

\label{secsub:categories}
\textbf{Persona Categories} To help organize the vast space of personas, we adopt the approach of \citet{lee-etal-2022-personachatgen} who introduced persona categories.
Persona categories are groups of persona attributes that describe the same semantic feature of the user.
In our work, we associate each persona category with a corresponding query that can be answered with all persona attributes in that category.
For example, job and family situation are persona categories, and corresponding queries might be ``What is your occupation?'', and ``Do you have a family?''.

\textbf{Persona Attribute Structure} Persona attributes can overlap. For instance, the attribute "I introduced my kids to scuba diving at a young age" overlaps with the attribute "My eldest son goes to elementary school", since both include the "parenthood" feature of the user.
Moreover, some persona attributes form a hierarchy, and some persona attributes are specific cases of other attributes.

\label{category_exists} \textbf{User Profile} We define a user profile as a set of persona attributes that can be used to describe a user.
For a realistic user, the persona attributes describing a user profile should not contradict each other, and be consistent. 
An arbitrary persona attribute set 
$U \subset \Omega$
is a consistent set of persona attribute if, and only if:
$$
\forall \pi_1 \in U,  \nexists \Pi_2 \subset U  : (\Pi_2 \neq \emptyset ) \land (\Pi_2 \rightarrow \neg \pi_1)
$$
% $\nexists \pi_1, \pi_2 \in U :  (\pi_1 \rightarrow \neg \pi_2~) \lor (\pi_2  \rightarrow \neg \pi_1)$

\textbf{Persona-based Conversation}  A persona-based conversation $D$ contains utterances such that at least one persona attribute from each user profile can be inferred from it.
For example, the persona attribute "I am a parent" can be inferred from the utterance "I just dropped off my son at school".
A persona-based conversation model is a generative model that takes a pair of user profiles ($U_1$, $U_2$) as input, and returns a persona-based dialogue $D$ between these two users.

\begin{figure}
    \centering
    \includegraphics[width=0.45\textwidth]{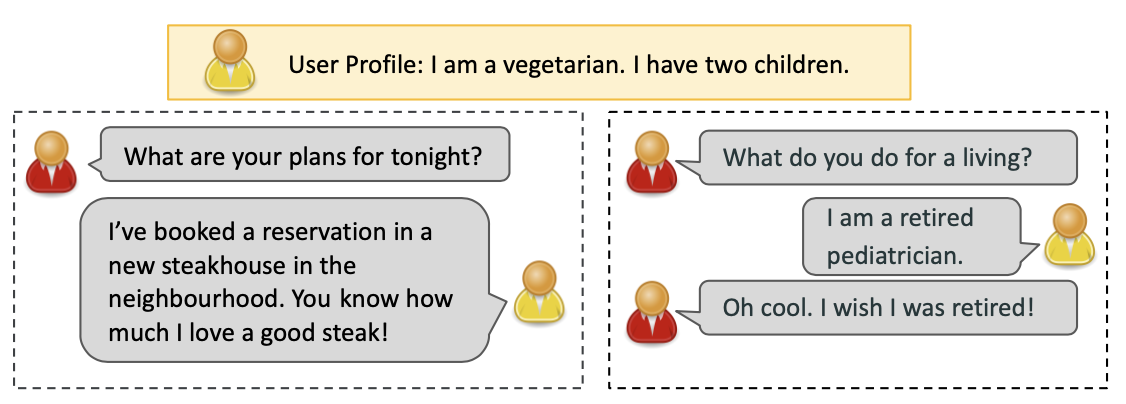}
    % \begin{minipage}{0.45 \textwidth}
    \caption{\small Unfaithful Conversation (Left): 
    Loving steak is negatively correlated with the persona attribute "I am a vegetarian".
    Faithful Conversation (Right): It introduces no information that contradicts or weakens the user's profile.
    }
    \label{fig:faithful_exp}
    % \end{minipage}
    \vspace{-5mm}
\end{figure}

\textbf{Faithfulness} 
One crucial quality for a persona-based conversation is that it should align with the user profile.
Inspired by \cite{daheim2023elastic} which introduces dialogue system faithfulness to the knowledge contained in relevant documents, we specify the criterion of \emph{faithfulness} to characterize the alignment between the utterances of a user in a persona-based conversation and their profile.
The faithfulness criterion enforces the constraint that the utterances of a user should not decrease the likelihood of their persona.
%not explicitly or implicitly contradict any of their persona statements.
This criterion assumes the existence of both a prior probability of persona attributes, and an inference model for determining the probability of persona attributes conditioned on utterances.
%We define the persona faithfulness criterion for conversations. 
%We leverage Natural Language Inference (NLI) to formalize the faithfulness criteria.
Let $M$ be such an inference model, ($U_1$, $U_2$) a pair of user profiles, and $D$ a persona-based conversation between them. 
To be a faithful conversation based on $M$, $D$ should not contain any contradicting evidence to the persona attributes of the speakers: passing the conversation $D$ as input to the inference model $M$ should not reduce the inference probability of persona attributes in either of the user profiles $U_1$ or $U_2$.
In other words, the probability of any persona attribute in the user profiles based on conversation $D$ should not be less than the probability of that persona attribute without any assumptions.
Formally, we call a conversation $D$ faithful with respect to the user profiles $U_1$ and $U_2$, and inference model $M$ if the following condition holds:
% \begin{center}
 $\forall \pi \in U_1 \cup~U_2: P_M(\pi|D) \geq P_M(\pi)$.   
% \end{center}
Where $P_M(\pi|D)$ indicates the probability that $M$ infers the persona $\pi$ given conversation $D$.
We show examples of faithful, and unfaithful conversations in Figure \ref{fig:faithful_exp}.

\section{Method}
\begin{figure}
    \centering
    \includegraphics[width=0.4\textwidth]{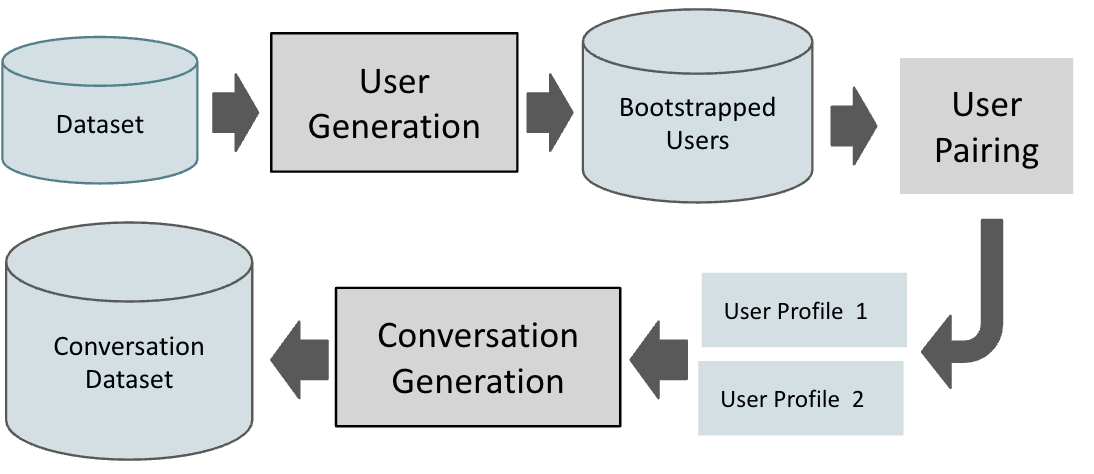}
    \caption{\small Dataset Augmentation Pipeline}
    \label{fig:arch}
    \vspace{-5mm}
\end{figure}

In this section, we introduce our method to generate persona-based conversations.
We create such conversations with minimum human input, starting from an initial dataset.
Our process consists of three steps, as shown in Figure \ref{fig:arch}: user generation, user pairing, and conversation generation. 
The first component augments a set of seed persona attributes $\Pi_0$ into an expanded set of persona attributes $\Pi_e$, from which it creates user profiles.
The second component pairs user profiles as interlocutors of a conversation. 
The third and final component uses an iterative process to generate high-quality conversations among user profile pairs. 
We detail each of these components below.
% Now, we describe each component:
% 
\subsection{User Generation}
The User Generation component is split into two sub-components:
\begin{enumerate}
    \item Persona Expansion
    \item User Profile Construction
\end{enumerate}
We bootstrap seed persona attributes by using various prompts \cite{brown2020language} to generate new persona attributes in the Persona Expansion step (Refer to Appendix \ref{app:persona_expand} for more details on the prompts used).
We then create new user profiles by iteratively selecting random user persona attributes from the expanded persona attributes.
We employ a Natural Language Inference (NLI) model to ensure the consistency of the constructed user profiles.

% We use a  Natural Language Inference (NLI) model to ensure the consistency of the constructed user profiles, removing any contradiction in them.
% The NLI model is a T5-based \cite{Raffel2019ExploringTL} model trained on TRUE \cite{honovich-etal-2022-true-evaluating} dataset.
% Figure \ref{fig:prof} in the appendix describes this process.
% set of features that measure specificity: we can use ontology, named entities 
\subsubsection{Persona Expansion}
\label{sec:pe}
We propose an unsupervised method to augment a set of seed persona attributes $\Pi_0$ into a super-set $\Pi_e$.
Unlike previous approaches \cite{lee-etal-2022-personachatgen}, our method is independent of human knowledge or intervention, making it capable of creating specialized personas in new domains. 
We proceed in two steps: query induction, and persona bootstrapping. 
In the query induction phase, we identify persona categories in $\Pi_0$, along with associated queries.
We then expand these queries into a set $Q$ that also covers unobserved persona categories.
The persona bootstrapping step leverages the category-based query set $Q$, and the initial persona attribute seed set $\Pi_0$ to generate new persona attributes.
Both of these steps are based on the bootstrapping technique \cite{yarowsky1995unsupervised}, and involve prompting an LLM.
We provide a detailed description of these two steps in the following.

% Ideally, the expanded set $\Pi_e$ becomes as large as the universe of potential personas $\Omega$. 
% In the query induction step, we create a query set $Q$ based on the distribution of personas in  $\Pi_0$, such that each query in $Q$ enforces the LLM to expand the persona set on a specific persona category. 

% Distribution shift and diversity and capability of generating more and work in more limited space- few shot cases

\textbf{Query Induction}
As described in Section \ref{secsub:categories}, each persona attribute belongs to at least one persona category, and each category is associated with a corresponding query that can be answered with persona attributes in that category. The query induction process initially identifies the queries associated with persona categories in $\Pi_0 $.
It then bootstraps queries by feeding them to a prompted LLM to create more queries that are associated with unobserved categories, ultimately creating a query set $Q$.
Including queries associated with unobserved persona categories facilitates the creation of a more diverse set of personas, and increases the scale of augmentation.

The query induction relies on the following assumption:

\textbf{Assumption }\emph{Let $\mathcal{M}$ be an LLM, and let $\Gamma$ be the set of all queries associated with all persona categories. If two persona attributes $\pi_1$ and $ \pi_2$ belong to the same persona category, then there exists a query $q^{\mathcal{M}} \in \Gamma$ such that $\pi_1$ and $\pi_2$ are $\mathcal{M}$'s output to $q^{\mathcal{M}}$.}

\begin{figure}
    \centering
    \includegraphics[width=0.42\textwidth]{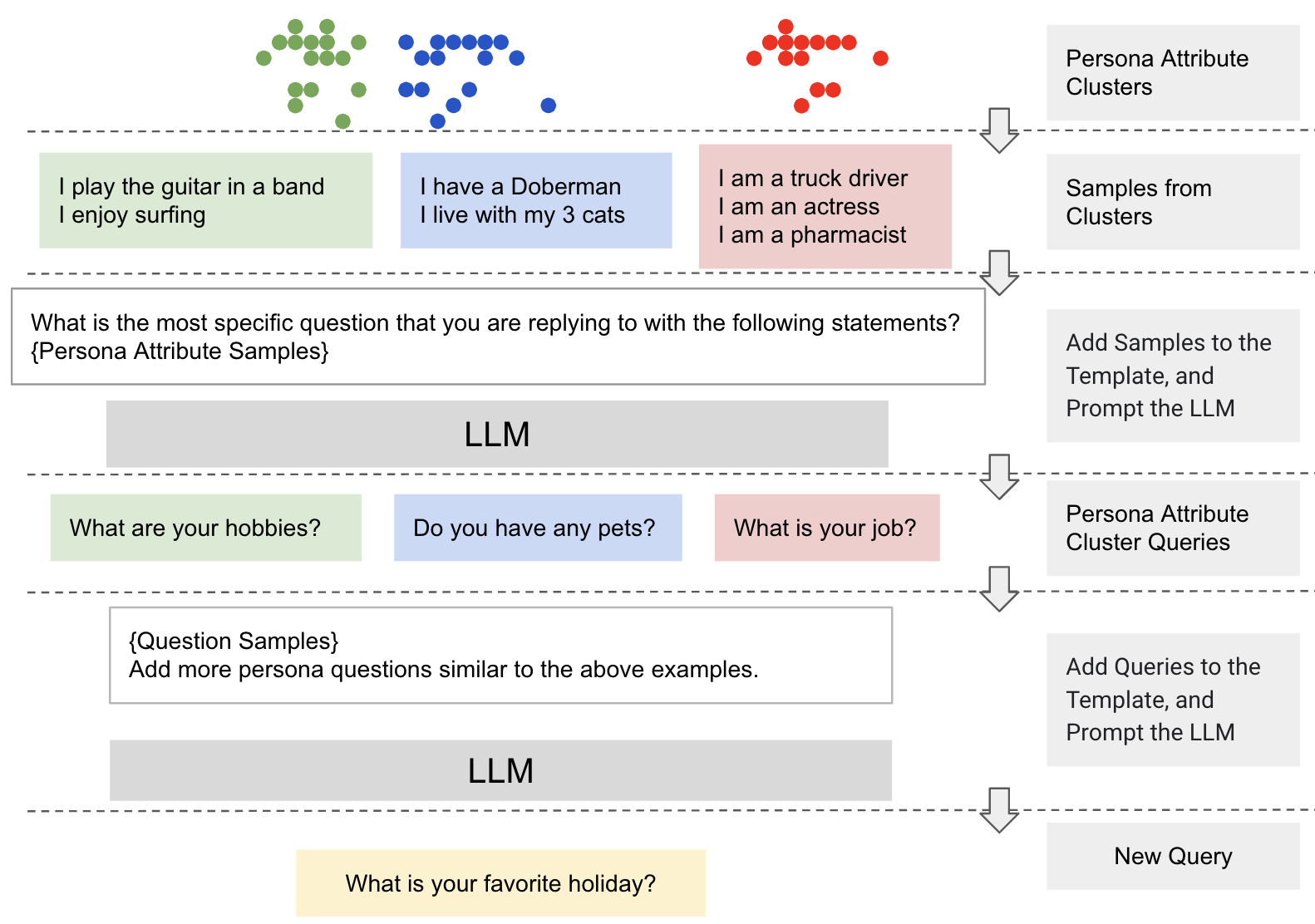}
    \caption{\small Query Induction Steps}
    \label{q_ind}
    \vspace{-4mm}
\end{figure}

The persona attributes "I am a doctor" and "I am a truck driver", for instance, both belong to the "job" category, leading to the query "What is your job?". 
We use an agglomerative clustering method to identify the persona categories in $\Pi_0$.
Let $C$ be an arbitrary persona cluster in $\Pi_0$.
To generate a query for $C$, we select a random subset of persona attributes in $C$, and create a prompt using these samples.
We employ this strategy to generate queries for all the clusters identified in $\Pi_0$, and create a set of queries, which we refer to as $Q_0$.
Details on the clustering, query induction, together with examples of clusters, persona attributes, and induced queries are available in Appendix \ref{app:persona_expand}.
We come up with queries for new, unobserved persona categories by bootstrapping the queries in $Q_0$:
starting from $Q = Q_0$, we iteratively sample a set of queries from $Q$, and create a prompt by concatenating them.
We then prompt the LLM to generate a new query, and add it to the query set $Q$, as shown in Figure \ref{q_ind}.
We generated a total of $|Q| = 188$ queries.
This set of category-specific queries $Q$ is later used to guide the LLM to generate new persona attributes from the specified category.
Thus, higher values of $|Q|$ result in greater diversity within the expanded persona attribute set.
% Thus, the higher value of $|Q|$ results in a greater diversity of the expanded persona set.

\textbf{Persona Bootstrapping}
We use the persona attribute seed set $\Pi_0$ and category-specific queries $Q$ to generate new persona attributes through a bootstrapping process.
We initialize $\Pi$ to $\Pi_0$.
At every iteration, we randomly select a subset of persona attributes from $\Pi$, and create a set of prompts as follows:
we first concatenate a set of persona attributes $s$.
For every query $q \in Q$, we then combine the concatenated samples $s$, and the query $q$ to create a category-specific persona prompt.
This prompt guides the LLM to generate a persona attribute for that persona category.
The set of prompts obtained from this process is 
$\{s  q | q \in Q \}$.
We only add a new persona attribute to the set if its BERT embeddings \cite{Devlin2019BERTPO} are not too close from existing ones, so as to prevent the addition of duplicates.

Each of these prompts is then fed to the LLM to create a new persona attribute, which is subsequently added to the set of persona attributes $\Pi$ for the next iteration.
We continue this iterative process until we have generated a total of 5k persona attributes.
Figure \ref{fig:p_bootstrap} illustrates the persona bootstrapping process.
Table \ref{tab:all_templates} in the appendix contains the prompt template used in this component.
% together to contribute as the few shot samples in the prompts and
\begin{figure}
    \centering
    \includegraphics[width=0.45\textwidth]{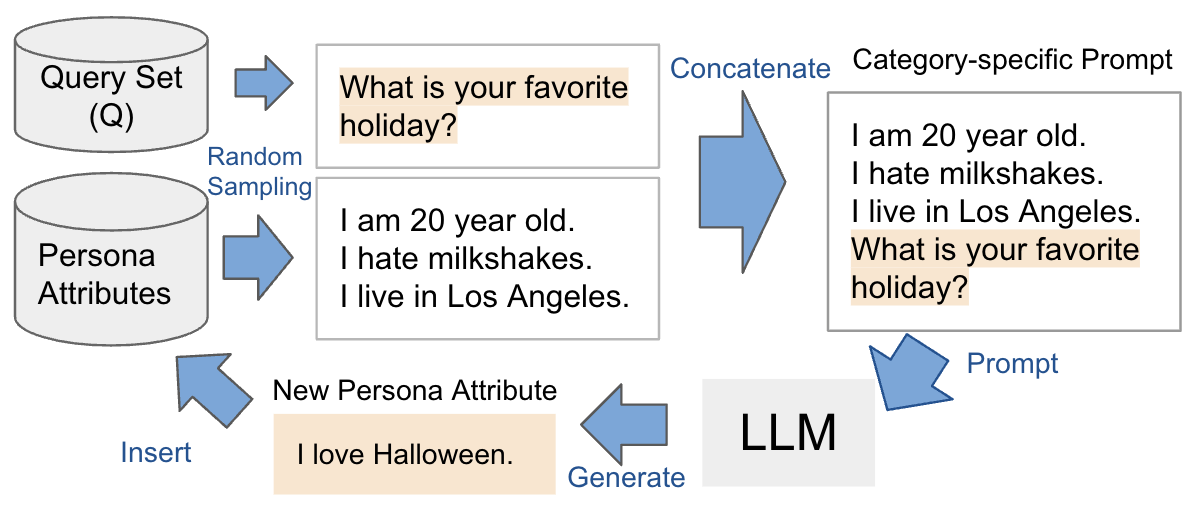}
    \begin{minipage}{0.45 \textwidth}
    \caption{\small Query-based Persona Bootstrapping Process}
    \label{fig:p_bootstrap}
    \end{minipage}
    \vspace{-7mm}
\end{figure}

\subsubsection{User Profile Construction}
We build user profiles incrementally by sampling persona attributes from $\Pi_e$, and adding the eligible ones.
A persona attribute is eligible if it adheres to the criteria of consistency and non-redundancy.
In other words, it should not contradict any attribute already in the user profile, and it should not be inferred by other persona attribute.
We assess the consistency and redundancy of user profiles by leveraging an NLI model, and persona attribute clustering, respectively.
The NLI model we employ is based on T5 \cite{Raffel2019ExploringTL}, and has been trained on the TRUE dataset \cite{honovich-etal-2022-true-evaluating}.

We create a user profile $U$ by iteratively selecting a random candidate persona attribute $\pi' \in \Pi_e$.
We use the NLI model to assess whether $\pi'$ contradicts any persona attribute in the profile.
This is determined by the condition: $\forall \pi \in U: (\pi' \not \rightarrow \neg \pi) \wedge (\pi \not \rightarrow \neg \pi')$, where $\rightarrow$ is an inference.
Additionally, we evaluate the similarity of $\pi'$ to the persona attributes in $U$ to prevent the addition of redundant attributes. 
% We also check the similarity of $\pi'$ to the personas in $U$ to avoid adding repetitive personas. 
We add $\pi'$ to $U$ if it meets the consistency and non-redundancy criteria.
We repeat this process until the user profile contains 5 persona attributes.
Please refer to Appendix \ref{subsect:usr_gen} for more details on the user profile construction.
% Figure \ref{fig:prof} in the appendix \ref{subsect:usr_gen} illustrates this process.

\subsection{User Pairing}
In this component, we identify potential pairs of users for conversations. 
As the conversations are persona-based, we hypothesize that they will be more engaging if the users' personas exhibit more commonalities.
We assign a similarity score to every pair of user profiles $(U_1, U_2)$, indicating their semantic similarity.
We leverage BERT to represent the user profiles. The similarity between $U_1$ and $U_2$ is defined as:
$
|\{(\pi_1, \pi_2)| \pi_1 \in U_1, \pi_2 \in U_2, \exists c: \pi_1, \pi_2 \in c\}|
$
Where $c$ is a persona attributes cluster.
The semantic similarity is quantified by the number of common persona categories in the user profiles.
We pair $U_1$ and $U_2$ if their similarity exceeds a threshold of 2.

\subsection{Conversation Generation}
\begin{figure}
    \centering
    % \begin{small}
    % \resizebox{0.5 \textwidth}{!}{
    \includegraphics[width=0.47\textwidth]{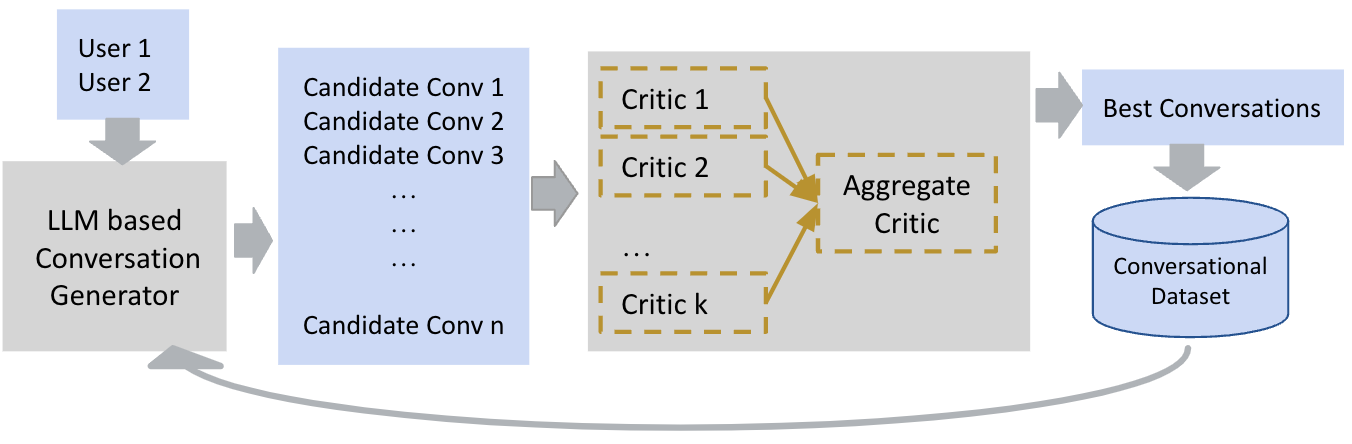}
    \caption{\small The Generator-Critic Architecture for Conversation Generation}
    % }
    \label{fig:get_crq}
    % \end{small}
    \vspace{-6mm}
\end{figure}

Our Conversation Generation component is similar to a general-purpose dataset generation framework that generates data samples, and refines them based on a set of predefined criteria, which we refer to as \emph{policies} \cite{madaan2023selfrefine}.
The flexibility in the choice of policies for data generation allows us to emphasize different objectives.
Once the active policies are selected, this component generates new data samples using a few input samples. 
The input to our Conversation Generation framework consists of a set of paired user profiles, a few samples of user profiles along with a persona-based conversation between them, and conversation quality metrics as policies.
We follow a Generator-Critic architecture, and iteratively create the dataset following the steps shown in Figure \ref{fig:get_crq}: \\
\textbf{Step 1} The Generator outputs candidate conversations between persona pairs using a few initial conversation samples. \\
\textbf{Step 2} The Critic evaluates the candidate conversations based on the predetermined policies, and selects the best candidate conversations. \\
\textbf{Step 3} The best candidate conversations are added to the dataset for the next iteration of generation.\\
This iterative process of selecting the top candidates and adding them to the dataset gradually improves the performance of the Generator. 

Without any loss of generality, we implement both the Generator and the Critic based on LLMs.
Specifically, the Generator prompts an LLM to create candidate conversations, while the Critic prompts an LLM to evaluate the quality of the generated conversations.

We provide more details on the Generator, Critic, and the policies we used.

The \textbf{Generator} outputs conversations for pairs of users $(U_1, U_2)$ by prompting an LLM \cite{brown2020language, wei2023chainofthought}. 
At each iteration, it randomly selects 5 samples from an initial set of conversations, each containing a pair of user profiles and a dialogue among them.
It feeds these samples to a template that instructs the LLM to generate a series of candidate conversations for the given user pair. 
The template, and a sample generated conversation are available in Table \ref{tab:all_templates}, and Table \ref{tab:sample_conv} in the appendix.

The \textbf{Critic} selects the best generated conversations to fine-tune the Generator.
A conversation is deemed high-quality if it complies with the policies of the Critic.
Given the multifaceted nature of the conversation evaluations, we use a Mixture of Experts (MoE) approach. 
Each expert evaluates the conversation based on a specific policy.
In this paper, we incorporate three types of experts, each with distinct criteria:
general conversation quality, persona faithfulness, and toxicity.
Collectively, these experts select the best generated conversations (the single best in our experiments).
We describe each type of expert, and the collective decision-making process below.

\textbf{General Conversation Quality} experts assess conversation quality using the \textbf{Fine-grained Evaluation of Dialog (FED)} metrics introduced in \cite{Mehri2020UnsupervisedEO}. 
These experts use verbalized forms of the policies from FED as prompts.
For instance, the "conversation depth quality expert" transforms the "depth policy" from FED into a prompt like "Which conversation is a deeper conversation between user 1 and user 2?".
Our system instructs the LLM to compare each pair of candidate conversations based on these policies, resulting in pairwise comparisons. 
The list of policies and their baseline performance are presented in Table \ref{tab:critque_perf} in Appendix \ref{app:fed_stat}.

The \textbf{Faithfulness} expert ensures the consistency of the generated conversations with the user profiles. 
It uses an LLM to identify instances of unfaithful conversations.
The faithfulness prompt provides the LLM with explicit instructions, user profiles, and human-curated examples of unfaithful conversations.
% , ensuring fidelity to the intended personas.

The \textbf{Toxicity} expert detects any conversation that exhibits harmful traits, including bias and hate.
% and eliminate them from candidate conversations. \\

The Critic filters unfaithful and toxic conversations out.
It then selects the best conversations using a majority vote among the General Conversation Quality experts. 
The selected instances are added to the dataset for the next iteration of the Generator.
\section{Evaluation}
We evaluate different aspects of our dataset generation framework, and the resulting dataset - referred to as \ourdataset{} - which is created using an instruction fine-tuned LLM with 24 billion parameters \cite{chung2022scaling}.
We compare \ourdataset{} (\ourdatasetabbr{}) against the widely used \personachat{} (\personachatabbr{}) dataset across different dimensions.
We begin by evaluating the quality of the personas we generate.
We then evaluate \ourdatasetabbr{} using both automatic metrics, and human assessment.
We analyze other aspects of \ourdatasetabbr{}, such as toxicity and diversity in appendices \ref{app:toxic} and \ref{app:div}.

\subsection{Evaluation of the Expanded Personas}
\label{sec:peq}
We evaluate our persona expansion module on two seed datasets: Wikipedia, and \personachat{}.
The Wikipedia personas are created by crawling the 1,000 most active contributors\footnote {
\url{https://en.wikipedia.org/wiki/Wikipedia:List_of_Wikipedians_by_number_of_edits}}, and extracting user boxes from their pages.
We expand both datasets using our framework, and evaluate the expanded persona attribute sets using automatic metrics.
Table \ref{tab:persona_q2} compares the original persona sets to the expanded ones on a few dimensions.
We observe that our persona expansion increases the number of persona attributes in \ourdatasetabbr{} by $119\%$, while maintaining the original persona categories and expanding them by $71\%$ compared to the persona attributes in \personachatabbr{}.
Moreover, the lengths of the new generated persona attributes are $107\%$ longer in \ourdatasetabbr{}, indicating that the new personas exhibit greater detail and specificity.
We observe a similar trend when applying our persona expansion to the Wikipedia persona set, with a $108\%$ increase in the number of persona attributes, a $140\%$ increase in persona categories, and a $45\%$ growth in persona attribute lengths.
This demonstrates the effectiveness of our method in expanding and diversifying persona sets.
% Table \ref{tab:persona_clusters} includes examples of new personas.
\begin{table}
    \centering
    \begin{small} 
    \resizebox{0.50\textwidth}{!}{
    \begin{tabular}{c|cc|cc}
    \toprule
    Dataset  & \personachat & \ourdataset  & Wikipedia & Wikipedia+  \\ 
    \midrule
    % & -Chat & -Chat+ & pedia & pedia+ \\ \hline
    $\#$ Persona Attributes & 4,723 & \textbf{10,371} & 8768 & \textbf{18,293}\\
    $\#$ Clusters & 323 & \textbf{553}  & 408 & \textbf{986}\\
    \small Inter-cluster Dist & 0.836 & 0.863  &0.816 & 0.85 \\
    % Distance & & & & & &\\
    AVG length & 7.65 & $\textbf{15.9}^*$ & 10.45 & $\textbf{15.2}^*$ \\
    % Perplexity & 118.53 & 46.75 & 1016 & 137.2\\
    % \hline
    % \hline
    % AVG \# turns & 7.4 & 11.4 & & & &\\
    \bottomrule
    \end{tabular}
    }
    \caption{ \small Evaluation of the expanded persona sets. The numbers with $^*$ indicate the metric value of the newly generated persona attributes to contrast with the initial set. 
    % C-Dist indicates the inter-cluster distance metric. 
    % Last row of is average number of turns per user in \personachat~ vs \ourdataset.
    }
    
    \label{tab:persona_q2}
    \end{small}
    \vspace{-5mm}
\end{table}
\subsection{Next Utterance Prediction}
\label{sec:nup}
A persona-based conversation reflects the speaker's persona explicitly or implicitly.
Therefore, we expect the inclusion of information about speaker personas to enhance the performance of next utterance prediction models in such conversations.
In this experiment, we assess the impact of incorporating speaker personas as prior information on both ranking, and generative - Transformer based \cite{Vaswani2017AttentionIA} - next utterance prediction models.
We create a subset of \ourdatasetabbr{} containing conversations among user pairs included in \personachatabbr{} for a fair comparison.
\begin{table*}
    \centering
    \begin{small}
    \begin{tabular}{c|c|ccc|ccc}
    \toprule
        &&  \multicolumn{3}{c|}{\personachat} & \multicolumn{3}{c}{\ourdataset}\\
        Method & Metric &  None   &  Persona & $\%$ Change & None  & Persona &  $\%$ Change \\ \midrule
        IR Baseline & hit@1 & 18.69 & 36.86 & +97 & 19.37 (19.92) &  \textbf{39.6} (26.23)&  \textbf{+104} (+31)\\
        % Profile Memory & & .2338 &  & & \\  
        % StarSpace  & & & & & \\
        \multirow{1}{*}{Transformer (Ranker)}
         & hit@1 & 14.24 & \textbf{19.21} & \textbf{+35} &9.71 (64.24)  & 11.74 (68.82) & +21 (+7)\\
        %  & MRR & 0.303 & 0.362 & +19 & 0.248 (0.763) & 0.270 (0.798) & +8 (+4)\\
        % (Ranker) & & & &  \\
        \midrule
        % \multirow{2}{*}{Seq2Seq} & hit@1 & .0489 & .0497 & .0709 &  .0649\\
        %  & ppl & 246.5 & 246.5 & 50.63 & 51.99\\
        \multirow{4}{*}{Transformer (Generator)}
         & hit@1 & \textbf{8.54}  & 6.78 & -20 & 6.89 (41.32) & 6.66 (37.35)&  \textbf{-3} (-9)\\
         &  Perplexity & \textbf{122.5} &  173.3 & +41 & 1032 (5.24) & 1126 (5.73) & \textbf{+9} (+9)\\
        & BLUE & \textbf{0.120} & 0.094 &-21 & 0.097 (0.289) & 0.083 (0.251) & \textbf{-14} (-13)\\
        & ROUGE &\textbf{0.141}  &0.113 &  -24 & 0.123 (0.348)  & 0.107 (0.309) & \textbf{-13} (-11)\\
        % \midrule
        %  \multirow{4}{*}{LMEDR\cite{Chen2023LearningTM}}
        %  & hit@1 &   &  &  & & &  \\
        %  &  Perplexity &  &  &  &  &  & \\
        % & BLUE &  &  &  &  &  &\\
        % & ROUGE & & & &  &  & \\
        
        % LMEDR & & .219 & &\\
        % LIC & & .177 & & \\
        % P2BOT &  & .197 & & \\
        \bottomrule
    \end{tabular}
    % \end{small}
    \caption{\small 
    Results of the next utterance prediction experiment. Performance of the trained model on the test split of \personachat{} is represented by the numbers in the table, while the numbers in parentheses indicate results for the test split of \ourdataset{}.
    }
    \label{tab:next_utr}
    \end{small}
    \vspace{-4mm}
\end{table*}
We observe (Table \ref{tab:next_utr}) that the performance of ranking models increases when personas are given to the models as input for both datasets. Specifically, the Transformer (Ranker) model, known for its ability to capture conversational complexity, exhibits higher performance in \ourdatasetabbr{} when evaluated on the \ourdatasetabbr{} test set compared to the \personachatabbr{} test set.
However, it demonstrates relatively weaker performance when trained on the \personachatabbr{}. This implies that \ourdatasetabbr{} contains more intricate and coherent conversations. 

The Transformer (Ranker) trained on \ourdatasetabbr{} achieves a hit@1 of 64.24 on \ourdatasetabbr{} test, $350\%$ higher than \personachatabbr{} (14.24). This suggests that the Transformer model can more accurately predict the next utterance in \ourdatasetabbr{}, pointing to a greater coherency in conversations. 

The performance of the Information Retrieval (IR) Baseline model is slightly higher for \ourdatasetabbr{}: it rises by $31\%$ when conditioned on user personas, which is lower than $97\%$ improvement in \personachatabbr{}.
A key contributing factor for the performance improvement of the retrieval-based model (IR Baseline) on \personachatabbr{} given the personas, is the participants' tendency to copy persona words in the conversations, whereas in \ourdatasetabbr{} the personas are more implicitly reflected in the conversations.
The implicit reflection of personas in \ourdatasetabbr{},  makes the task more challenging for word based retrieval models, necessitating reasoning that goes beyond word level.
However, when the model is trained on \ourdatasetabbr{} and tested on \personachatabbr{}, the improvement is as high as when the model is trained on \personachatabbr{}, i.e. $104\%$ compared to $97\%$. 

The performance of generative models is low for this task since these models are not trained with the ranking objective.
However, the performance difference while the models are conditioned on personas is lower for the model trained on \ourdatasetabbr{}, with a $20\%$ drop for the model trained on \personachatabbr{} against $3\%$ drop in the model trained on \ourdatasetabbr{}.
The increase in perplexity is $9\%$ in \ourdatasetabbr{} compared to $41\%$ in \personachatabbr{}.
The lower rate of perplexity increase and performance drop of the model given user personas as input highlights the higher alignment of conversations with personas in \ourdatasetabbr{}. 

We also evaluate the performance of the next utterance prediction models when given no user, one user, and both user personas. The results suggest a higher degree of bidirectionality in \ourdatasetabbr{}.
We refer the reader to the Appendix \ref{app:nup} for more details.

\subsection{Human Evaluation}
We compare the quality of the conversations generated by our framework against those in \personachat{}.
We randomly select 200 conversations from \personachatabbr{}, together with their corresponding user pairs, and use our method to generate conversations among the same users.
We start by following \cite{Gehrmann2019GLTRSD} in running a human experiment to try and detect AI-generated content.
We conduct a Turing test where we present pairs of conversations to humans, and ask them to identify the synthetically generated one. 
This test is carried out on the generated conversations at the end of each iteration of creating \ourdatasetabbr{}.
We repeat the test for conversations generated for new persona pairs, which we refer to as iteration $3^*$, i.e. we pair each of these conversations with a random conversation from \personachatabbr{}.
For a robust evaluation, every pair of conversations is annotated by 3 human evaluators, and the majority vote is used as the final annotation.
Details of this test are available in Appendix \ref{app:tur_test}.
The results of this experiment can be found in Table \ref{tab:human_eval_iter}. 
We observe that the losing rate of \ourdatasetabbr{} is reduced by $48\%$ 
from \ourdatasetabbr{} Iter 1 to \ourdatasetabbr{} Iter 3, and dropped below the rate of $10\%$.
Interestingly, $91\%$ of the conversations in \ourdatasetabbr{}, which are synthetically generated, are judged as human-like as the conversations generated by humans. Moreover, conversations generated for new personas (Iteration $3^*$) are deemed artificial in only $8.04\%$ of cases, showing that \ourdatasetabbr{} is more realistic than \personachatabbr{}.

\label{exp:h_faith}
We also evaluate the faithfulness of the generated conversations.
For each conversation, we provide annotators with a faithfulness annotation task including the speakers' persona attributes and distractor persona attribute options as shown in Figure \ref{fig:h_fact}.
We evaluate faithfulness during 3 iterations of conversation generation for the selected 200 user pairs, and the annotators evaluate the generated conversations for each pair in every iteration.
The results show that, while improving the Turing test results, faithfulness of conversations are consistently higher than $75\%$ with at most $3\%$ variation in between iterations, indicating high faithfulness in all iterations.

% Finally, we evaluate the effect of the LLM size on quality of the generated dataset in our framework. We create a version of \ourdataset{} using the LLM2 with 540 billion parameters. Table \ref{tab:human_eval_iter} shows the human evaluation of the generated dataset with the smaller LLM but in multiple iterations and the generated dataset in one iteration with the LLM2.
% We observe that while in the first iteration of dataset generation, the dataset generated by the larger model performs $5\%$ better in the Turing test, after two iterations those conversations outperform in the test. This shows the capability of our framework to create conversations with competitive quality but lower computations using iterative data generation. 
Finally, we assess the impact of LLM size on the quality of the generated dataset within our framework.
We create a variant of \ourdatasetabbr{} using an LLM with 540 billion parameters (LLM2).
Table \ref{tab:human_eval_iter} presents human evaluations comparing the smaller LLM in multiple iterations to a single-iteration approach with LLM2.
The larger model exhibits a 5\% advantage in the Turing test over the first iteration of dataset generation over the smaller model.
After two iterations, however, the multi-iteration approach outperforms the first iteration of the bigger model, showing our framework's capacity for cost-effective, high-quality conversation generation.
\begin{table}
    \centering
    \begin{small}
    \resizebox{0.47\textwidth}{!}{
    \begin{tabular}{c|c|c|c||c}
        % \hline
        \toprule
        Conversation Source &  Lose & Win & Tie  & Faithful \\ 
        % \cline{1-4}
        \midrule
        \cmidrule(r){1-4}
        % \cmidrule{r}{5-5}
        \ourdatasetabbr{} Iter 1 & 17.2 & 30.1 & 52.68 & 78.5\\ %.364  \\
        \ourdatasetabbr{} Iter 2 &  18.5 & \textbf{49} & 32.5 & 80.5 \\ %.427  \\
        \ourdatasetabbr{} Iter 3 &  \textbf{8.8} &  35.23 & \textbf{55.95} & 
        76.6 \\
        
        \cmidrule(r){1-4} 

        % \midrule 

        \ourdatasetabbr{} Iter 3* & 8.04  & 32.66 & 59.29 & N/A\\
        \ourdatasetabbr{} (LLM2) & 11.5 & 39 & 49.5 & N/A\\ \bottomrule
    \end{tabular}}
    \caption{\small 
    Turing Test on 200 Generated Conversations per Iteration: \ourdataset{} Outcomes Against \personachat{}. 
    % The columns "Win", "Lose", and "Tie" represent the percentage of \ourdataset{} outcomes compared to \personachat{}.
    % Faithfulness is measured by precision of extracted personas. Humanness column is the percentage of conversations that were not detected as synthetically generated. 
    % Note that last iteration of \ourdataset is the weak evaluation of the conversations based on the extended persona set.
    }
    \label{tab:human_eval_iter}
    \end{small}
    \vspace{-6mm}
\end{table}
% \subsection{Implicity of Personas}
% A natural feature of persona based human conversations is the presence paraphrased or implicit persona descriptions rather than explicit persona descriptions. For example for a given persona "I am a male", the phrases like '' or '' are more likely to appear in the conversation rather than the \pegah{vanilla} phrasing of the persona. 
% In this experiment, we evaluate this natural feature of persona based conversations i.e. not directly repeating the personas. 
% For every conversation, we compute the word overlap between the speaker personas and each conversation. 
% \pegah{multiple choice of persona selection. 
% Compare the performance with model trained on Persona chat. !Hakim: warning. how are you using training data??}
% To ensure of fair comparison, we select a subset of the conversations in our dataset (called $X$), which have common personas with persona chat. In other words, for any conversation $C=(User 1, User 2, D) \in X$, there exists a conversation in persona chat $C' = (User 1, User 2, D')$ with the same speakers and human written conversation.  
% We created a human experiment to compare the overall quality of the generated conversations with persona chat for any pair of user personas in persona chat, we showed the generated conversation in $X$ and personaChat conversation and asked human annotators to select the best conversation. Every pair of conversation we annotated by three people.
% Table \pegah{Y} shows the results of this experiment. 
% \section{Analysis}
\section{Related Work}
Large Language Models (LLMs) have been used for data augmentation \cite{shin-etal-2021-constrained}, generation \cite{kim2023soda, Dong2023RAFTRR}, and evaluation \cite{Zhang2019BERTScoreET, Liu2023GEvalNE}.
One of the earliest works in this area \cite{AnabyTavor2019NotED} used LLMs to create a large text dataset from a small, labeled one.
This idea was followed by \cite{Wang2021TowardsZL, Schick2021GeneratingDW} which leveraged LLMs to create datasets without any human data.
\cite{Kumar2020DataAU} evaluated the performance of different LLMs on the data augmentation task.
Several conversational dataset generation methods focused on the structure of the conversational data \cite{Dai2022DialogIT, Leszczynski2023GeneratingSD, Abbasiantaeb2023LetTL}. \cite{mehri-etal-2022-lad} illustrated how Large Language Models (LLMs) can effectively generate synthetic training data for task-oriented dialogue models.

% showed the potency of LLMs as data proxies for task oriented conversations. 

Persona-based conversations have been a popular research topic in NLP \cite{Liu2022PersonaBasedCA}.
One of the earliest works in this area is \personachat{}, by \cite{Zhang2018PersonalizingDA}, which proposed the Persona-Chat dataset and evaluation metrics that have become a benchmark for persona-based conversation generation \cite{mazare-etal-2018-training}.
Many subsequent works have used this dataset to train and evaluate their models, including DialoGPT \cite{zhang2020dialogpt}, BlenderBot \cite{Shuster2022BlenderBot3A}, and PersonaChatGen \cite{lee-etal-2022-personachatgen}.
PersonaChatGen automated the process of creating persona based conversations of \personachat{} using LLMs.
A challenge in generating synthetic datasets is to ensure the quality of the conversation including data faithfulness, fidelity, diversity, and consistency \cite{Li2016APN, lee2023_llm_data_creation, veselovsky2023generating, Zhuo2023RedTC,Wang2023DecodingTrustAC, Mndler2023SelfcontradictoryHO}.
Several works have focused on creating and using high quality training datasets \cite{welleck2019dialogue}, and creating quality filtering components to their conversation dataset generation \cite{lewkowycz2022solving}.
Evaluation of the resulting conversational datasets is also challenging \cite{xu2021goldfish}.
\cite{wang2023rethinking} recently introduced the paradigm of interactive evaluation of conversations with LLMs.

% Data faithfulness and fidelity refers to the extent to which generated conversations resemble those that occur in real life. Several works have addressed this issue in the context of persona-based conversation generation, including Towards Faithful Personalized Response Selection in Retrieval Based Dialog Systems\pegah{\cite{}}. Another work in this area is 
% \cite{Lin2022TeachingMT}. 

\section{Conclusion and Future Work}
We developed a novel framework for generating high-quality persona-based conversations using LLMs, resulting in the creation of \ourdataset{}, comprising 20k conversations. We hope this dataset will support future endeavors in developing persona-aware conversational agents, including the generation of domain-specific multi-session conversations for specialized, task-oriented interactions. While we focused on a persona-based dataset generation task, our Generator-Critic approach can be generalized to other use cases, such as generating other specialized datasets, etc.

\section*{Limitations}
In this paper, we define an iterative process over LLMs to generate a dataset.
Our method requires computational resources, and access to an LLM.
The quality of the dataset is bounded by the LLM, since the quality critics are also using the same LLM, and we leave the iterative improvement of our critics as future work.
The main limitation of this data generation framework is the inability to generate realistic conversations that do not have high quality, since we assume that both parties are fluent, that the conversation flow is perfectly consistent, and there is no unexpected event (e.g. an interruption by another person, connection loss, etc.) in the middle of the conversation.  
Another limitation of our method is the difficulty of incorporating less tangible persona traits, such as a sense of humor, or user attributes that require multiple conversation sessions to be reflected. 

% ACL 2023 requires all submissions to have a section titled ``Limitations'', for discussing the limitations of the paper as a complement to the discussion of strengths in the main text. This section should occur after the conclusion, but before the references. It will not count towards the page limit.
% The discussion of limitations is mandatory. Papers without a limitation section will be desk-rejected without review.

% While we are open to different types of limitations, just mentioning that a set of results have been shown for English only probably does not reflect what we expect. 
% Mentioning that the method works mostly for languages with limited morphology, like English, is a much better alternative.
% In addition, limitations such as low scalability to long text, the requirement of large GPU resources, or other things that inspire crucial further investigation are welcome.

\section*{Ethics Statement}
The approach of generating datasets based on some desired objective might be used to create harmful datasets, and train malicious models based on them, such as a biased dataset, or a hateful speech one \cite{Hartvigsen2022ToxiGenAL}.
On the other hand, these datasets and models can be used as filters in application tasks. 

We used Amazon Mechanical Turk in our human experiments, and followed that platform's guidelines to protect the rights of human raters.
The participation was voluntary, and the raters were informed of their rights at the beginning of the study.
The platform implemented security measures to protect them, and prevent the disclosure of any Personal Identifiable Information about them.
Furthermore, we offered higher than minimum standard wage compensation to avoid any exploitative practices.

To avoid having any toxic conversation in the final dataset, we also used several tools to remove any potentially toxic conversation.
Details about these tools, and example removed samples are available in Appendix \ref{app:toxic}. 

% Scientific work published at ACL 2023 must comply with the ACL Ethics Policy.\footnote{\url{https://www.aclweb.org/portal/content/acl-code-ethics}} We encourage all authors to include an explicit ethics statement on the broader impact of the work, or other ethical considerations after the conclusion but before the references. The ethics statement will not count toward the page limit (8 pages for long, 4 pages for short papers).

\section*{Acknowledgements}
The authors would like to thank Kian Ahrabian, Eric Boxer, Luke Friedman, Iñaki Iturrate, Kathy Meir-Hellstern, Filip Radlinski, and Kexuan Sun for their valuable comments on this manuscript.

% This document has been adapted by Jordan Boyd-Graber, Naoaki Okazaki, Anna Rogers from the style files used for earlier ACL, EMNLP and NAACL proceedings, including those for
% EACL 2023 by Isabelle Augenstein and Andreas Vlachos,
% EMNLP 2022 by Yue Zhang, Ryan Cotterell and Lea Frermann,
% ACL 2020 by Steven Bethard, Ryan Cotterell and Rui Yan,
% ACL 2019 by Douwe Kiela and Ivan Vuli\'{c},
% NAACL 2019 by Stephanie Lukin and Alla Roskovskaya, 
% ACL 2018 by Shay Cohen, Kevin Gimpel, and Wei Lu, 
% NAACL 2018 by Margaret Mitchell and Stephanie Lukin,
% Bib\TeX{} suggestions for (NA)ACL 2017/2018 from Jason Eisner,
% ACL 2017 by Dan Gildea and Min-Yen Kan, NAACL 2017 by Margaret Mitchell,https://braintex.goog/project/637bfa2c5ee2590082180724 
% ACL 2012 by Maggie Li and Michael White, 
% ACL 2010 by Jing-Shin Chang and Philipp Koehn, 
% ACL 2008 by Johanna D. Moore, Simone Teufel, James Allan, and Sadaoki Furui, 
% ACL 2005 by Hwee Tou Ng and Kemal Oflazer, 
% ACL 2002 by Eugene Charniak and Dekang Lin, 
% and earlier ACL and EACL formats written by several people, including
% John Chen, Henry S. Thompson and Donald Walker.
% Additional elements were taken from the formatting instructions of the \emph{International Joint Conference on Artificial Intelligence} and the \emph{Conference on Computer Vision and Pattern Recognition}.

\bibliography{acl2023}
\bibliographystyle{acl_natbib}

\newpage
\newpage

\appendix

% \section{Appendix}

\label{sec:appendix}

\section{Dataset Generation Framework}
In this section, we provide more details on our synthetic dataset generation framework.
We created \ourdataset{}  using an LLM with 24 billion parameters.
We use top-k sampling with $k=40$ for decoding during generation, and set the temperature value to 0.7 in all components.
We give more details on user and conversation generation components in the following subsections.
\subsection{User Generation}
In our framework, the user generation component consists of two steps: expanding the persona attribute set, and creating realistic user profiles.
In this section we provide details on our framework for these two steps:
\paragraph{Persona Expansion}
\label{app:persona_expand}
As described in Section \ref{sec:pe}, the persona expansion step 
involves identifying persona categories in the initial persona attribute set $\Pi_0$, generating queries associated with those categories, and bootstrapping queries to create a query set $Q$. In our framework, we employ the Scikit-learn \cite{scikit-learn} implementation of an agglomerative clustering to identify persona categories following this clustering method:
we represent each persona using a BERT-based representation. 
Our clustering approach is bottom-up, starting with each persona attribute as an individual cluster.
At each step, we combine two clusters if their similarity exceeds a predetermined threshold of 0.1.
The similarity of two clusters is measured using inter-cluster average cosine similarity. The process continues until no pair of clusters is more similar than the threshold.

After identifying the clusters, we sample 3 instances of persona attributes for each cluster, and prompt the LLM using the template in shown in section \ref{q_ind} to construct an initial query set $Q_0$. 
We expand the query set $Q_0$ using bootstrapping.
At each step, we sample 5 instances from the available queries, and prompt the LLM using the template in Table \ref{tab:all_templates}.
We repeat this process for 100 steps. 
Examples of initial persona attributes, induced queries, bootstrapped queries, and bootstrapped persona attributes can be found in Table \ref{tab:persona_clusters}. The prompt templates used in this component are available in Table \ref{tab:all_templates}. 

\begin{figure}
    \centering
    \includegraphics[width=0.45\textwidth]{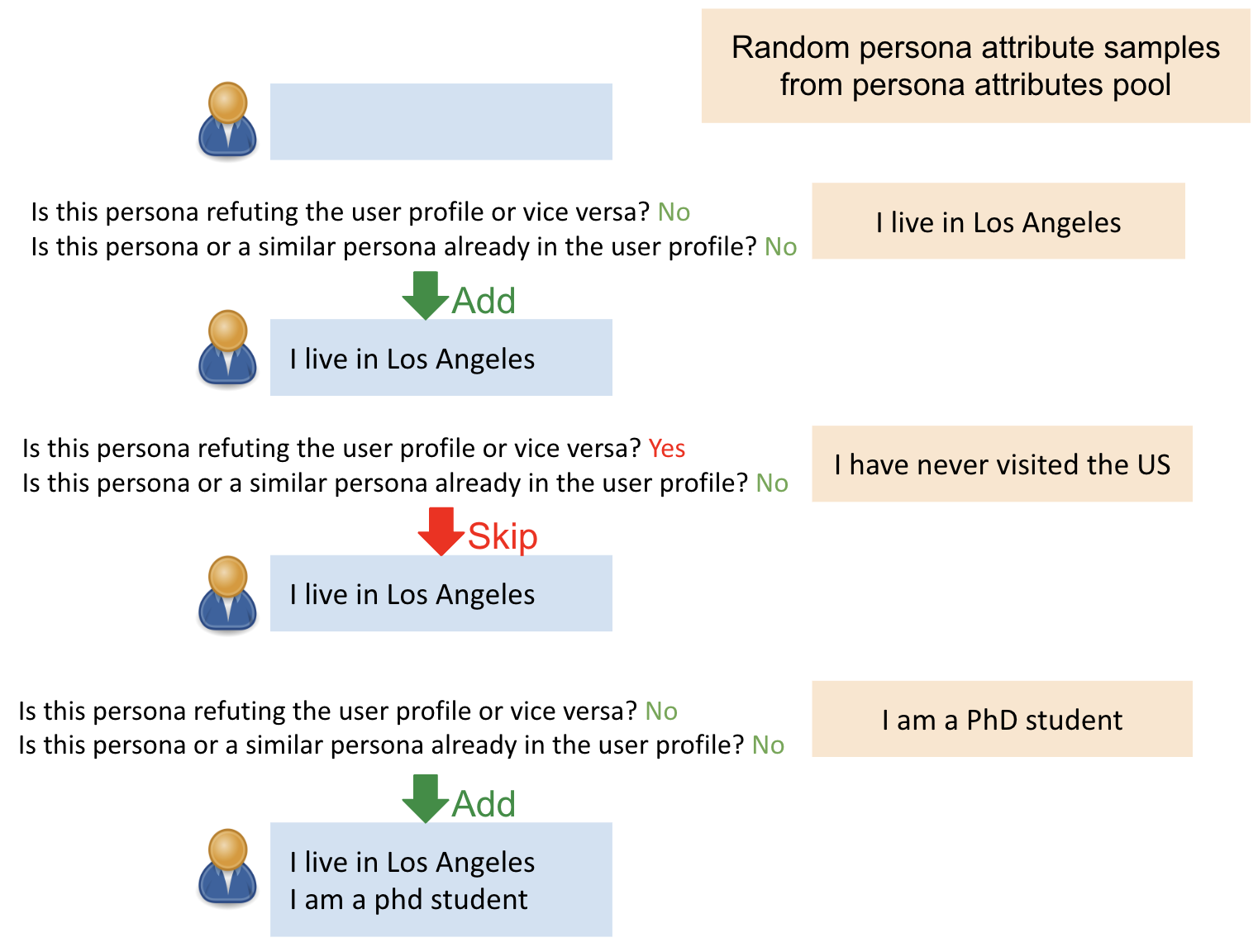}
    \caption{\small User Profile Construction Example}
    \label{fig:prof}
\end{figure}

\paragraph{User Profile Generation}
% \subsubsection{User Profile Generation}
\label{subsect:usr_gen}
We illustrate a sample user profile creation process in Figure \ref{fig:prof}.
As shown in the figure, at each iteration, a randomly selected persona attribute is checked for consistency and non-redundancy.

Let $\pi'$ be a randomly selected persona attribute in an iteration.
For the redundancy criteria, we use the BERT representation of persona attributes.
We compute the similarity of the new candidate persona attribute $\pi'$ with every persona attribute in the user profile.
If it is more than a threshold (0.9 in these experiments) similar to an attribute in the user profile, $\pi'$ is deemed as redundant and will not be added to the user profile.
We use the cosine similarities of the BERT representations of the persona attributes.

For the consistency criteria, we use the NLI model to verify the consistency of this persona attribute with the user profile.
For every persona attribute in the current user profile $\pi$, we prompt the LLM to create the negated persona attribute $\neg \pi$.
Then, we query the NLI model to check whether $\neg \pi$ is inferred by $\pi'$ or $\neg \pi'$ is inferred by $\pi$.
If either of these cases is inferred, then the selected persona attribute is not consistent with the user profile, and not added to the profile.
\begin{table*}
    \centering
    % \begin{small}
    % \resizebox{0.99\textwidth}{!}{
    \begin{tabular}{p{1cm}|p{1.2cm}|p{4.8cm}|p{7cm}}
    \toprule
         Dataset & Persona Source & Query & Example Persona Attribute \\ \midrule
         \parbox[h]{9mm}{\multirow{6}{*}{\rotatebox[origin=c]{90}{\personachat}}}
         & \multirow{3}{*}{Human} & What is your job? & I am a pharmacist.\\
         &  & Where do you live? & I live close to the coast.\\
         & & Do you have any pets? & I have a doberman.\\
        \cmidrule{2-4}
        & \multirow{3}{*}{LLM} & What are your talents? & I am a great listener.\\ 
        &  & What is your hair color? & My hair is auburn.\\
        &  & What is your favorite song? & I like the song "Leather and Lace".\\ 
        \bottomrule
        \toprule
        \parbox[t]{2mm}{\multirow{6}{*}{\rotatebox[origin=c]{90}{Wikipedia}}}
        &  \multirow{3}{*}{Human} & What are your hobbies? & I spend WAY too much time on Wikipedia.\\ 
        &  & What is your view on the metric system? & I find the metric system to be a logical and efficient way to measure things.\\
        \cmidrule{2-4}
        & \multirow{4}{*}{LLM} & What is the name of the first album you ever purchased? & My first album was The Miseducation of Lauryn Hill \\
        &  & What are you interested in? & I'm looking to learn new recipes and improve my cooking skills. \\

        \bottomrule
    \end{tabular}
    % }
    
    \caption{
    Persona Categories and Induced Queries Using Our Framework.
    Queries are generated by the Large Language Model (LLM). 
    Queries for personas with the "LLM" as source, are generated through bootstrapping, while those with "human" as source are generated by sampling persona categories and prompting the LLM.
    Personas with "human" as the source are authored by humans, while "LLM" rows represent personas generated using our framework.
    }
    \label{tab:persona_clusters}
\end{table*}

\subsection{Conversation Generation}

\paragraph{LLM-based Critic}
\label{app:fed_stat}
In our framework, the critic is implemented by prompting an LLM. We included a mixture of experts approach in the critic, where each expert prompts the LLM to assess a specific policy in the candidate conversations. Our framework includes a set of experts to control the general conversation quality. We evaluate the performance of these experts using a baseline dataset. The baseline dataset for this experiment is FED which consists of 125 human-annotated instances evaluated at the conversation level. We pair the conversations and evaluate the experts based on the number of correctly ranked pairs. As shown in Table \ref{tab:critque_perf}, we observe that these experts are more than $80\%$ accurate in distinguishing the better conversation within the pairs. The template for the verbalized form of these experts used in our framework can be found  in Table \ref{tab:all_templates}.

% \centering
 \begin{table}
    % \centering
    \begin{center}
    \begin{small}
    \begin{tabular}{c|c}
    \toprule
        Policy &  Performance \\ \midrule
        % Consistency   & \\
        % Fluency   & \\
        Depth  & 0.84  \\
        Coherency  & 0.96  \\
        Consistency  & 0.92 \\
        Diversity   & 0.92 \\
        % Flexibility  & \\
      %  Understandable   & .53\\
        Likable   & 0.88 \\
   %     Informative   & .63\\
    %    Inquisitive   & .40\\
        % Engaging & \\
        % \hline
        % Bias & & \\
        \bottomrule
    \end{tabular}
    \caption{\small 
    List of FED Experts for Persona-Based Conversation Generation Critic. Performance is measured by the number of correctly compared conversation pairs in FED baseline based on the given policy.}
    \label{tab:critque_perf}
    \end{small}
    \end{center}
\end{table}
% \end{center}

\begin{table*}
    \centering
    \begin{tabular}{p{3.5cm}|p{13cm}}
        \toprule
        \small Component & \small Template \\
        \midrule
        \small Query Induction & 
        \small
        What is the most specific question that you are replying to with the following statements? 
        \newline
        \{persona-category-sample-1\} \newline
        \{persona-category-sample-2\} \newline
        \{persona-category-sample-3\}
        \\
        \midrule
        \small Query Bootstrapping & 
        \small
        \{cluster-query-1\} \newline
        ... \newline
        \{cluster-query-5\} \newline
        Add more persona questions similar to the above examples. \\
        \midrule
        \small Persona Bootstrapping &  
        \small
        Imagine you are a person with the following persona. \newline
        \{random-persona-attribute-1\} \newline
        ... \newline
        \{random-persona-attribute-5\} \newline
        \{query\}. Answer with only one short sentence that starts with 'I' or 'My'. Do not repeat the given persona.
        \\ \midrule
        \small FED Expert &  
        \small
        Which one of Conversation 1 and Conversation 2 between two users \{policy\}? Why? 
        \newline
        Conversation 1: \{conv-1\}
        \newline
        Conversation 2: \{conv-2\}
        \\ \midrule
        \small
        Toxicity Expert & 
        \small Is this conversation toxic? Why? 
        \newline
        Conversation: \{conv\}
         \\ \midrule
        \small Conversation Generation &  
        \small
        Here, we list the profiles of two users, user 1 and user 2, followed by an interesting and natural conversation between user 1 and user 2, which implicitly reflects their user profiles. \newline
        User 1 Profile: \{conversation1-user-1\} 
        \newline
        User 2 Profile: \{conversation1-user-2\}
        \newline
        Conversation: \{conversation-1\}
        \newline
        ...
        \newline
        User 1 Profile: \{conversation-5-user-1\} 
        \newline
        User 2 Profile: \{conversation-5-user-2\}
        \newline
        Conversation: \{conversation-5\}
        \newline
        Give me more examples like this. The conversation must be more than 5 turns and less than 8 turns. The conversation must be natural, and not direct copies of their profiles. \newline
        User 1 Profile: \{user-1\} \newline
        User 2 Profile: \{user-2\}
        \\
        \midrule
       \small
        Faithfulness Expert & 
        \small Given user 1 and user 2's profiles respectively, does the following conversation between the two users contradict either of their profiles? Why? \newline
        User 1 Profile: \{user-1\}\newline
        User 2 Profile: \{user-2\}\newline
        Conversation: \{conv-1\} \newline
        Response: \{explanation\}
        \\ \bottomrule
    \end{tabular}
    \caption{\small
    Prompting Templates for Large Language Models of Different Components in Our Framework. Variables enclosed in \{\} are filled when the template is populated.}
    \label{tab:all_templates}
\end{table*}

We also included a toxicity expert and a persona faithfulness expert in the critic. The prompt templates used in these experts are available in Table \ref{tab:all_templates}. The persona faithfulness leverages in-context- learning capability of LLMs. It includes a few human-curated examples of faithful and unfaithful conversations in the instruction prompt. Refer to Table \ref{tab:faithful_expert_few_shot} for examples of faithful and unfaithful conversations used in the instruction prompt. 
\begin{table*}
    \centering 
    \resizebox{\textwidth}{!}{
    \begin{tabular}{p{5cm}|p{9cm}|p{3cm}}
    \toprule
    User Profiles & Conversations & Faithfulness Explanation\\
    \midrule
    \small
    \textbf{User 1}: \newline
    One of my favorite hobbies is gaming. \newline
    I currently work in an office.   \newline
    I enjoy running \newline

    \textbf{User 2}: \newline
    I am applying for publishing jobs. \newline
    My mother is a librarian. \newline
    The only autographs I've ever wanted are from authors.\newline
    I used to wear glasses as a child. \newline
    My favorite color is green. \newline
    
        & 
        \small
        
        User 1: Do you like video games? 
        
        User 2: No, I do not like video games.
        
        User 1: What do you do for a living?
        
        User 2: I am a librarian.
        
        User 1: I guess that is why you like to read a lot. 
        
        User 2: Yes, I like to read books. 
        
        User 1: What is your favorite color? Mine is green.
        
        User 2: That is my favorite color too. 
        
        User 1: That is cool.
        
        User 2: Do you have any pets? 
        
        User 1: No, I do not have any pets. 
        
        User 2: I have a dog and a cat.
        
        User 1: That is cool.
        
        User 2: Do you have any hobbies?.
        
        User 1: Yes, I like to write.
        
        User 2: What do you write about?
        
        User 1: I write about my life.
        
        User 2: That is cool. 
        
        User 1: Thanks!
        % \\ \cmidrule{2-2}
        & \small
        No, the conversation does not contradict the profiles.
        Neither user 1 nor user 2 says anything that contradicts their respective profiles. 
        Although the conversation does not cover everything in their profiles, that is ok.
        \\ \midrule
        \midrule
        \small
        \textbf{User 1:} \newline
        I would like to move some place warmer. \newline
        I m an out of work farmhand. \newline
        I like watching reality television shows. \newline
        I live in Kansas.\newline

        \textbf{User 2:} \newline
        I am studying to be a dentist. \newline
        My favorite team is the raptors. \newline
        My favorite band is rise against. \newline
        I usually go to bed late. \newline
        I like sports.
        & 
        \small
        User 1: It's so cold in Kansas, I want to go some place warmer.
        
        User 2: I am studying dentistry.
        
        User 1: I am an out of work farmhand, I love reality tv.

        User 2: I like the raptors and rise against, I go to bed late.

        User 1: It is very hard to find work here, I might need to go to the city.

        User 2: I don't like sports , but I like the raptors .
        
        User 1: I would like to work as a model, but I have no experience.
        
        User 2: I am applying for a job as a receptionist.
        
        User 1: I am very jealous, I hope you get it.
    % \\ \cmidrule{2-2}
    & \small
    Yes, the conversation contradicts user 2's profile. In the profile, user 2 says they like sports, however, in the conversation, user 2 says they don't like sports.
    \\
    \midrule \midrule
   \small 
    \textbf{User 1}: 
    
    My family is from Kazakhstan.
    
I can speak Spanish, English, and Kazakh.

I like to go hiking on the weekends.

I play jazz piano in a band.

I am a barista.
\newline

\textbf{User 2:} 

I am a vegan.

I love game of thrones.

I like soccer.

My brother used to be in the navy.
&
\small
User 1: Hi, what do you do for a living?

User 2: I am a barista.

User 1: Awesome, me too! Where are you from?

User 2: My family is from Kazakhstan, but I was born in Uruguay.

User 1: Cool . What do you like to do in your free time?

User 2: I love to go hiking and play jazz piano.

User 1: Wow, that sounds really nice. I like to watch game of thrones and play soccer.

User 2: I have never seen game of thrones, but I love soccer.

User 1: You should watch it, it is really good.

User 2: I will check it out.
% \cmidrule{2-2}
& 
\small
Yes, the conversation contradicts user 2's profile. In the profile, user 2 says they love game of thrones, but in the conversation user 2 says they have never seen it.
\\ \midrule
\midrule
\small
\textbf{User 1:} \newline
My favorite book is David Copperfield. \newline
I have a twin brother. \newline
I am really short. \newline
I love to bake cakes. \newline
I have won tap dance competitions. \newline

\textbf{User 2:} \newline
I enjoy spending a lot of money on my hobbies. \newline
I love photography. \newline
I like to use candles instead of lights. \newline
I do not like waking up early.
&
\small
User 1: Hi there! Do you like to read?

User 2: I like to read. I like to go to the library.

User 1: Oh, cool! What is your favorite book?

User 2: I really like David Copperfield. I have a twin brother and I am really short.

User 1: Are you good at baking?

User 2: I can bake cakes. I have won tap dance competitions.

User 1: Do you like photography?

User 2: I love taking pictures. I also like to use candles instead of lights.

User 1: Do you like to wake up early?

User 2: I like to sleep in. I do not like waking up early.

User 1: Do you like to spend money?

User 2: I love spending money on my hobbies.
% \\ \cmidrule{2-2}
&
\small
No, the conversation does not contradict either user profiles. 
Although user 2 says they are really short, which is in user 1's profile, it doesn't mean user 2 cannot be really short too. 

\\
\bottomrule
    \end{tabular}
    }
    \caption{\small 
    Examples of Faithful and Unfaithful Conversations for Prompting the Persona Faithfulness Expert in Our Framework. The `Faithfulness explanation' columns are used to fill the `explanation' variable in the faithfulness expert template. }
    \label{tab:faithful_expert_few_shot}
\end{table*}

\section{\ourdataset}
\ourdataset{} is made of 20k conversations, with an average of 11.8 turns per user for each.
An example \ourdataset{} conversation can be found in Table \ref{tab:sample_conv}. 
We compare \ourdataset{} to \personachat{} across different dimensions.
We first assess the characteristics of \ourdatasetabbr{} using various automatic evaluators, i.e. evaluators which do not require human effort.
We then conduct a human evaluation experiment on a subset of \ourdatasetabbr{}.
\begin{table*}
    % \begin{small}
    % \resizebox{\textwidth}{!}{
    
    \begin{tabular}{p{4cm}|p{11cm}}
    \toprule
    User Profiles & Conversation \\ 
    \midrule
    \textbf{User 1} 
    
    I am male. 
    
    My weakness are sweets.  
    
    I am unsociable. 
    
    I am a jack of all, master of none. \newline

    \textbf{User 2}
    
    My favorite music is rock and roll.
    
    I love to travel.
    
    I am from Russia.
    
    I enjoy skiing.
    
    I go to school for business. 
    
    & 
User 1: Hello

User 2: Hello, what's your name? 

User 1: My name is [name], and you? 

User 2: My name is [name], I'm from Russia.

User 1: Nice to meet you, [name]. I'm from the United States.

User 2: It's nice to meet you too, [name]. What are you interested in?

User 1: I'm interested in a lot of things, but my main interests are music and travel.

User 2: I love music too! What kind of music do you like?

User 1: I like a lot of different kinds of music, but my favorite is rock and roll.

User 2: Me too! I love classic rock and roll. Who are your favorite bands?

User 1: My favorite bands are The Beatles, The Rolling Stones, and Led Zeppelin.

User 2: Those are all great bands! I love them too.

User 1: What about you, what are your favorite bands?

User 2: My favorite bands are The Doors, Pink Floyd, and Queen.

User 1: Those are all great bands too! I love them too.

User 2: So, what do you like to do for fun?

User 1: I like to go to concerts, play video games, and watch movies. 

User 2: I love going to concerts too! I also like to go skiing and play tennis.

User 1: I love skiing too! I haven't played tennis in a while, but I used to play a lot. 

User 2: We should go skiing together sometime! 

User 1: That would be fun! I'm free this weekend if you are.

User 2: I'm free this weekend too! We should go skiing.

User 1: Great! I'll text you the details. 
\\
        \bottomrule
    \end{tabular}
    % }
 
    \caption{\small Sample Conversation from \ourdataset{}. This conversation was synthesized from user profiles in \personachat{}}
    \label{tab:sample_conv}
\end{table*}

\subsection{Automatic Evaluation}
We conduct a comprehensive analysis and evaluation of \ourdatasetabbr{} across different dimensions and compare it against \personachatabbr{}.
We start by analyzing the toxicity and diversity of \ourdatasetabbr{} using off the shelf tools. Then, we elaborate on the experiments which assess the efficacy of \ourdatasetabbr{} used as the dataset for the next utterance prediction and the profile extraction tasks.
Finally, we evaluate the quality of \ourdatasetabbr{} conversations using LLM-based evaluation methods. 

\paragraph{Toxicity Analysis}
\label{app:toxic}
We analyze the toxicity of the generated conversations at the final iteration of \ourdatasetabbr{} using an online tool called Perspective\footnote{https://perspectiveapi.com/}. 
We reproduce the results of a detailed analysis of toxicity in \personachatabbr{} as well as in each iteration of our data generation framework while producing \ourdatasetabbr{} in Table \ref{tab:toxicity}.
\begin{table*}
    \centering
    \resizebox{\textwidth}{!}{
    \begin{tabular}{c|ccc|ccc}
    \toprule
         & & Toxicity & & & Profanity  & \\
         Confidence & weak(< .2) & medium(.2-.8) & strong(>.8) & weak(< .2) & medium(.2-.8) & strong(>.8) \\ \midrule
        % Dataset   & & & & & \\
        \personachatabbr & 10875 & 4448 & 53  & 10891 & 1676 & 57\\
        \ourdatasetabbr{} Iter 1 & 10902 & 1192 & 3 & 10903 & 340 & 3 \\
        \ourdatasetabbr{} Iter 2 & 10900 & 1096 & 1 & 10901 & 345 & 1 \\
        \ourdatasetabbr{} Iter 3 & 10902 & 1088 & 1 & 10902 & 376 & 0\\
        \bottomrule
    \end{tabular}
    }
    \caption{Frequency of Toxic Conversations in \personachat{} and \ourdataset{}}
    \label{tab:toxicity}
\end{table*}

We observe a notable reduction in the frequency of conversations deemed as strongly toxic or profane throughout the iterations of generating \ourdatasetabbr{}.
This reduction can be attributed to the built-in toxicity filter of the employed LLM.
While \personachatabbr{} contains more than 50 samples that are identified as strongly toxic, \ourdatasetabbr{} includes at most three toxic or profane conversations, which is significantly lower (at least 15 times less).
Interestingly, the fraction of conversations with medium profanity and toxicity in \ourdatasetabbr{} is 4 times less than the same type of conversations in \personachatabbr{} across all iterations.
We have removed any conversation that was marked as strongly toxic by this tool in the released dataset.
Samples of toxic conversations are provided in Table \ref{tab:toxic_samples}.
\begin{table*}
    \centering
    \begin{tabular}{p{3.5cm}|p{11.5cm}}
        \toprule
        \small Source & \small Conversation \\ \hline
        \small \personachat &  
        \small
        ... \newline
        User 1: I like bloody stuff. \newline
        User 2: It reminds me of the dark which makes me afraid of it. \newline
        User 1: You are a silly goose. 
        \\ \midrule
        \small \personachat & \small
        ... \newline
        User 2: Cool. Why do you say that? Because I am a red head? \newline
        User 1: No. Ikn. Why do you ask so many questions? Mr. Thomas is dumb. 
        \\ \midrule
        \small \ourdataset & 
        \small 
        User 1: I can imagine. What's your favorite part of the job? \newline
        User 2: I love working with my team and seeing our restaurant succeed. \newline
        User 1: That's great. What's your least favorite part of the job? \newline
        User2: My least favorite part is dealing with my boss. He's a real jerk. 
        \\ \bottomrule
    \end{tabular}
    \caption{\small Examples of Toxic Conversations. The first two examples are segments of conversations from \personachat{}. The final example is a segment from a toxic conversation in \ourdataset{}, which has been removed in the released dataset.}
    \label{tab:toxic_samples}
\end{table*}

\paragraph{Diversity Analysis}
\label{app:div}
We use hierarchical topic modeling \cite{blei2004hierarchical} to assess the topic diversity of \ourdatasetabbr{} and compare it to that of \personachatabbr{}.
For a fair comparison, we only compare conversations in \ourdatasetabbr{} with similar personas in \personachatabbr{}.
Table \ref{tab:div} displays the number of topics at each level of the topic tree, with the first level indicating the most general topic.
We observe similar topic diversity at the first level. In deeper levels, there is a slightly lower diversity in \ourdatasetabbr{}.
% \pegah{Add the percentage of change}
% \pegah{any justification?}
\begin{table}
    \centering
    \begin{tabular}{c|c|c}
        \toprule
        Topic Level &  \personachatabbr & \ourdatasetabbr\\  \midrule
        1 & 27 & 27 \\
        2 & 232 & 213 \\
        3 & 470 & 403 \\
        4 & 137 & 118 \\
        5 & 30 & 26 \\
        \bottomrule
    \end{tabular}
    \caption{\small Vertical Topic Diversity in Persona-based Datasets}
    \label{tab:div}
\end{table}

\paragraph{Next Utterance Prediction}
\label{app:nup}
We compare the performance of different models on the next utterance prediction task.
As discussed in Section \ref{sec:nup}, these models are expected to exhibit better performance in the next utterance prediction task when user personas are provided as prior information.
We evaluate ranking and generative models for response selection to assess this property.
We compare models trained on \ourdatasetabbr{} to the same models trained on \personachatabbr{}.
We use the implementations provided in \cite{miller2017parlai} for the following models:
\begin{itemize}
    \item 
        \textbf{IR Baseline} Given an utterance as a query, the IR baseline finds the most similar utterance in the training corpus using tf-idf.
        It defines the utterance after the most similar utterance as the candidate response, and then returns the most similar option to that candidate as the output.  
    \item 
    \textbf{Transformer-Ranker} The context of the conversation, as well as the candidate next utterances, are encoded using a BERT-based encoder. The most similar encoded candidate to the conversation context, as measured by a dot-product in their representation space, is selected as the output \cite{humeau2020polyencoders}.
    \item \textbf{Transformer-Generator}
This model is a sequence-to-sequence model \cite{Sutskever2014SequenceTS} which uses transformers as encoders and decoders. 
%  \begin{table*}
%      \begin{small}
%      \resizebox{\textwidth}{!}{
%      \begin{tabular}{c|c|cc|cc|cc|cc}
%      \toprule
%           & &  \multicolumn{2}{c|}{No Persona} & \multicolumn{2}{c|}{Self Persona} & \multicolumn{2}{c|}{Their Persona} & \multicolumn{2}{c}{Both Personas} \\
%          Method & Metric & \personachat & \ourdataset &  \personachat & \ourdataset  & \personachat & \ourdataset &  \personachat & \ourdataset \\ \midrule
%          IR baseline & hit@1 &0.1869 & 0.1861 & \textbf{0.3683} & 0.2596 & 0.1519 & 0.1882 & 0.3281 & 0.2493\\
%         %  StarSpace & & & & \\
%         %  Profile Memory & & & & \\
%          Transformer(Ranker) & hit@1 & 0.2513 & 0.7164 &  0.275 &0.6227 &0.1922 & 0.6988 & 0.2572 &0 .7214\\
%         %   & & & & & & & &&\\
%          \midrule
%          Transformer & hit@1 &0.0896 & 0.0526 & .08512 & 0.629 & 0.0873 & 0.053 &0.0813 & 0.051 \\
%          (Generator) & ppl & 65.57& 5.54 & 72.24 & 5.47 & 62.49& 5.4 & 64.07 & 5.405 \\
%          \bottomrule
%      \end{tabular}
%      }
%      \caption{\small Evaluation of Next Utterance Prediction Models Conditioned on Different Users' Personas as Input. The first column for each condition represents results on \personachat{}, while the second column shows results on \ourdataset{}.}
%      \label{tab:nxt}
%      \end{small}
%  \end{table*}
 
  \begin{table*}
     \begin{small}
     \resizebox{\textwidth}{!}{
     \begin{tabular}{c|c|cccc|cccc}
     \toprule
        & & \multicolumn{4}{c|}{\personachat} & \multicolumn{4}{c}{\ourdataset} \\
        % \multicolumn{2}{c|}{No Persona} & \multicolumn{2}{c|}{Self Persona} & \multicolumn{2}{c|}{Their Persona} & \multicolumn{2}{c}{Both Personas}
     
          Method & Metric &  No Persona & Self Persona & Their Persona & Both Personas &  No Persona & Self Persona & Their Persona & Both Personas\\
           \midrule

         IR baseline & hit@1 &0.1869 & 
         \textbf{0.3683} & 
         0.1519 & 
         0.3281 & 
        0.1861 & 
        \textbf{0.2596} & 
        0.1882 & 
         0.2493\\

        %  StarSpace & & & & \\
        %  Profile Memory & & & & \\
         Transformer(Ranker) & hit@1 & 
         0.2513 & 
         \textbf{0.275} &
         0.1922 & 
         0.2572 &
         0.7164 & 
         0.6227 &
         0.6988 & 
         \textbf{0.7214} \\
        %   & & & & & & & &&\\
         \midrule
         Transformer & hit@1 &
         \textbf{0.0896} & 
         0.08512 & 
         0.0873 & 
         0.0813 & 
         
         0.0526 & 
         \textbf{0.629} &
         0.053 &
         0.051 \\
         (Generator) & ppl & 
         65.57& 
         72.24 & 
         \textbf{62.49}& 
         64.07 & 
         
         5.54 &
         5.47 & 
         \textbf{5.4} &
         5.405 \\
         \bottomrule
     \end{tabular}
     }
     \caption{\small Evaluation of Next Utterance Prediction models conditioned on different user personas.}
     \label{tab:nxt}
     \end{small}
 \end{table*}
\end{itemize} 

We also evaluate the performance of the next utterance prediction models when given no user, one user, and both user personas.
The results of this experiment are available in  Table \ref{tab:nxt}.
We observe that the highest performance improvement for all models trained on \personachatabbr{} is when self-personas are given as input.
We do not observe such a pattern in \ourdatasetabbr{}.
% We observe that using both user personas as input result in highest improvement in the model performance in \ourdataset{} compared to giving only one user personas to the model. 
This indicates a higher degree of bidirectionality in \ourdatasetabbr{} conversations compared to those of \personachatabbr{}.

\paragraph{Profile Extraction}
A potential use-case of the \ourdatasetabbr{} dataset is training a model to predict user personas from a conversation.
This is only possible if the dataset is highly faithful, meaning that any persona attribute inferred from the conversation is in the user profile or compatible with the user profile.
In this context, a faithful conversation is expected to have high precision in the profile extraction task, while a conversation that highly reflects user personas is expected to have high recall in this task.

We evaluate the task of user profile extraction for conversations in \ourdatasetabbr{}, and compare the results against those of \personachatabbr{}.
We frame the task of profile extraction as a ranking task, using the utterances within the conversations as queries.
The goal is to rank a set of persona attribute options.
For each conversation, we include the speakers' persona attributes in the available options.
Additionally, we select 25 random user persona attributes from other speaker profiles within the dataset to serve as distractors. 
The input to the profile extraction is utterances from a single user as the speaker, while the output is a list of persona attribute options for a target user, which could be either user 1 or user 2.
The results of this experiment are presented in Table \ref{tab:profile_extraction}.
We observe that the performance of the profile extraction methods
is higher in \ourdatasetabbr{} in 3 of the 4 scenarios.
Interestingly, we observe that with both datasets, when the target and the speaker are different, the performance of profile extraction is greater compared to the cases when the target and speaker users are the same.
% \pegah{Include both precision and recall}

% \pegah{We also observe the trend of having higher accuracy of model when the persona targets and utterance are by the same person. }
\begin{table}
    \centering
    \begin{small}
    \begin{tabular}{cc|cc}
        \toprule
        & & \multicolumn{2}{c}{F-Score} \\
        Target & Speaker &  \personachatabbr & \ourdatasetabbr \\
        \midrule
        % user 1  & user 1  & .938 & .958 & .931 & .952\\
        % user 1  & user 2  & .896 & .929 & .907 & .934\\
        % user 2  & user 1  &  .896 & .874 & .874 & .903\\
        % user 2  & user 2  & .878 & .908 & .907 & .933\\
        user 1  & user 1  & 0.505  &    \textbf{0.574} \\
        user 1  & user 2  & \textbf{0.737} &     0.68\\
        user 2  & user 1  & 0.50     &  \textbf{0.57}   \\
        user 2  & user 2  & 0.456 &   \textbf{0.494} \\
        \bottomrule
    \end{tabular}
    \caption{
    Accuracy of Profile Extraction in Four Different Scenarios. The `Target' column represents the user profile to be extracted, while the `Speaker' column indicates the speaker of the turns given to the model as input.
    }
    \label{tab:profile_extraction}
    \end{small}
\end{table}

\paragraph{LLM-based Quality Evaluation}
\label{llmeval-sec}
We leverage LLM-based conversation quality evaluators from the literature to compare the quality of \ourdatasetabbr{} and \personachatabbr{}.
These evaluators rely on the human curated prompt templates for different metrics including consistency, fluency, etc.
We used these evaluators with minimum change in the original prompt templates.
These evaluators are:
\begin{itemize}
    \item \textbf{LLM-Eval} \cite{Lin2023LLMEvalUM} is a multi-dimensional automatic evaluation designed for conversations.
    It uses a human-curated prompt which describes evaluation dimensions, serving as a unified evaluation schema.
    This prompt evaluates the conversation across multiple dimensions (e.g. fluency) in a single model call.
    We show this unified schema in Table \ref{tab:autoprompts}.
    \item \textbf{GPT-Score} \cite{Fu2023GPTScoreEA} leverages emergent abilities of LLMs, i.e. zero-shot instructions, to score texts.
    It contains a prompt template, and for each quality criterion, populates the template with a human description of the criteria along with the valid score range for that criteria.
    Example prompts are provided in Table \ref{tab:autoprompts}. 
    \item \textbf{G-Eval} \cite{Liu2023GEvalNE} introduces a framework that employs LLMs with a chain-of-thought approach to assess the quality of natural language generated outputs.
    For any evaluation criteria, G-Eval prompts the LLM with the criterion's description, 
    prompting the model to generate the necessary evaluation steps.
    It then uses these steps to prompt the LLM to score given output for that criterion. It considers the probability of getting each permissible score as the output of the prompt, i.e., it considers the probability distribution of scores assigned by the LLM. The reported output is the expected value of the score distribution by the LLM.
    Table \ref{tab:autoprompts} includes an example prompt.
\end{itemize}

\begin{table*}
    \centering
    \begin{tabular}{p{1.8cm}|p{2cm}|p{11.2cm}}
        \toprule
        Evaluator & Metric & Prompt Template \\
        \midrule
        LLM-Eval & All & \small Human: The output should be formatted as a JSON instance that conforms to the JSON schema below. \newline

As an example, for the schema \{"properties":
\{"foo": \{"title": "Foo", "description": "a
list of strings", "type": "array", "items":
\{"type": "string"\}\}\}, "required": ["foo"]\}\}
the object \{"foo": ["bar", "baz"]\} is a
well-formatted instance of the schema.
The object \{"properties": \{"foo": ["bar",
"baz"]\}\} is not well-formatted. \newline

Here is the output schema:
\{"properties": \{"content": \{"title":
"Content", "description": "content score
in the range of 0 to 100", "type":
"integer"\}, "grammar": \{"title": "Grammar",
"description": "grammar score in the range
of 0 to 100", "type": "integer"\}, "relevance":
\{"title": "Relevance", "description":
"relevance score in the range of 0 to 100",
"type": "integer"\}, "appropriateness":
\{"title": "Appropriateness", "description":
"appropriateness score in the range of 0 to
100", "type": "integer"\}\}, "required":
["content", "grammar", "relevance",
"appropriateness"]\}
        \newline
        
Score the following dialogue generated on a continuous scale from \{score-min\} to \{score-max\}.\newline
Dialogue: \{dialogue\}

\\ 
        GPT-Score & Consistency &  \small Answer the question based on the conversation between two users.
        
        Question: Are the responses of
users consistent in the information they provide throughout the conversation? (a) Yes. (b) No.

Conversation:
\{dialogue\}
Answer:

        \\
        G-Eval & Coherence & \small
        You will be given a pair of user personas. You will then be given one conversation between this persona pair.

Your task is to rate the conversation on one metric.

Please make sure you read and understand these instructions carefully. Please keep this document open while reviewing, and refer to it as needed.\newline

Evaluation Criteria:\newline

Coherence (1-5) - the collective quality of all utterances. We align this dimension with the Document Understanding Conference (DUC) quality question of structure and coherence (\url{https://duc.nist.gov/duc2007/quality-questions.txt}), whereby "the conversation should be well-structured and well-organized. The conversation should not just be a heap of related information, but should build from utterance to a coherent body of conversation about a topic."\newline

Evaluation Steps:\newline

1. Read and understand the given conversation between the pair of user personas.\newline
2. Evaluate the conversation based on the coherence of the utterances.\newline
3. Rate the conversation on a scale of 1 to 5, with 5 being the highest coherence and 1 being the lowest coherence.\newline
4. Justify the rating by referring to specific aspects of the conversation that demonstrate its coherence or lack thereof.\newline

Example:\newline

Personas: \{personas\}

Conversation: \{dialogue\}\newline

Evaluation Form (scores ONLY): 

- Coherence:
        \\
 \midrule       
    \multirow{3}{*}{LLM-Faithfulness}     & Inference & \small
Instruction: Select User \{user\} persona attributes that are directly inferred from this conversation.
\\
 & Contradiction & \small
Instruction: Select User \{user\} persona attributes that strongly contradict this conversation.
\\
     \bottomrule
    \end{tabular}
    \caption{\small Prompt Templates in LLM-based Conversation Quality Evaluators. Variables enclosed in \{\} are filled when the template is populated.}
    \label{tab:autoprompts}

\end{table*}

Results of this evaluation are presented in Table \ref{tab:auto-eval}.
We observe that \ourdatasetabbr{} consistently outperforms \personachatabbr{} across all the dimensions we evaluate.
The superiority of \ourdatasetabbr{} is more prominent when using GPT-Score, for which each evaluated criterion shows an average improvement of at least 23 points.
\begin{table*}
    \centering
    \begin{tabular}{cc|cc||ccc}
        \toprule
         Evaluator & Criteria & \personachatabbr & \ourdatasetabbr & \ourdatasetabbr{} Iter 1 &  FED &  Faithfulness\\ 
         \midrule
         \multirow{4}{*}{LLM-Eval \cite{Lin2023LLMEvalUM} }
         & Content & 81.96 & 	\textbf{88.84}	& \textbf{88.71} & 	87.61 & 	88.67 \\
& Grammar& 87.12 & 	\textbf{93.64}	& \textbf{93.68} & 	93.09	& 93.56 \\
& Relevance & 86.82	& \textbf{94.16} & 	\textbf{93.81}	& 92.88 & 	93.79 \\
& Appropriateness& 86.99	& \textbf{95.84}	& 96.17 & 	95.68 &	\textbf{96.19} \\
        \midrule
        \multirow{6}{*}{GPT-Score \cite{Fu2023GPTScoreEA}} &
 Fluency & 67.04 &	\textbf{98.89} &	96.28 &	96.65 &	\textbf{97.83}\\
&Consistent & 3.47&	\textbf{64.25}&	\textbf{50.43}&	43.45&	48.69\\

&Coherent& 69.41&	\textbf{100}	& \textbf{100} &	98.99&	\textbf{100}\\
&Depth &5.40& \textbf{37.36} &	\textbf{29.30}&	19.40&	29.01\\
&Diversity & 72.98&	\textbf{96.42} &	94.02	& 92.79 &	\textbf{94.11}\\
&Likeable &
          36.53 &	\textbf{91.04} & \textbf{93.11} &	91.90 &	87.98 \\
         \midrule
         \multirow{5}{*}{G-Eval \cite{Liu2023GEvalNE}}
         & Relevance (1-5) & 2.288 & \textbf{2.992} &	2.986	& 2.941	& \textbf{2.99} \\
		& Fluency (1-3) & 1.928 &	\textbf{2.002} &	\textbf{2} &	1.998 &	1.999 \\
		& Consistent (1-5) & 1.736 & \textbf{2.651} &	\textbf{2.587} &	2.449 &	2.496 \\
		& Coherent (1-5) & 2.505 & \textbf{2.997} & 2.997 & 2.991 & \textbf{2.998} \\
		& Faithfulness (1-5) &1.754 &	\textbf{2.959} & \textbf{2.8801} & 2.79 &	2.868 \\

    \bottomrule 
    \end{tabular}
    \caption{Results of Automatic Evaluations of \ourdataset{} and \personachat{}. The "FED" column is the evaluation of the dataset generated without FED expert and the column "Faithfulness" is the evaluation results of the dataset generated without the faithfulness expert in the Critic.}
    \label{tab:auto-eval}
\end{table*}

\subsection{Human Evaluation}
\label{app:tur_test}
We run a human evaluation of the performance of our method via a crowdsourcing platform.
We conduct a Turing test, and a faithfulness study - both of which we describe in more details in the following subsections - at the end of every iteration of the generation of \ourdatasetabbr{}.

\begin{figure*}
    \centering
    \includegraphics[scale=0.73]{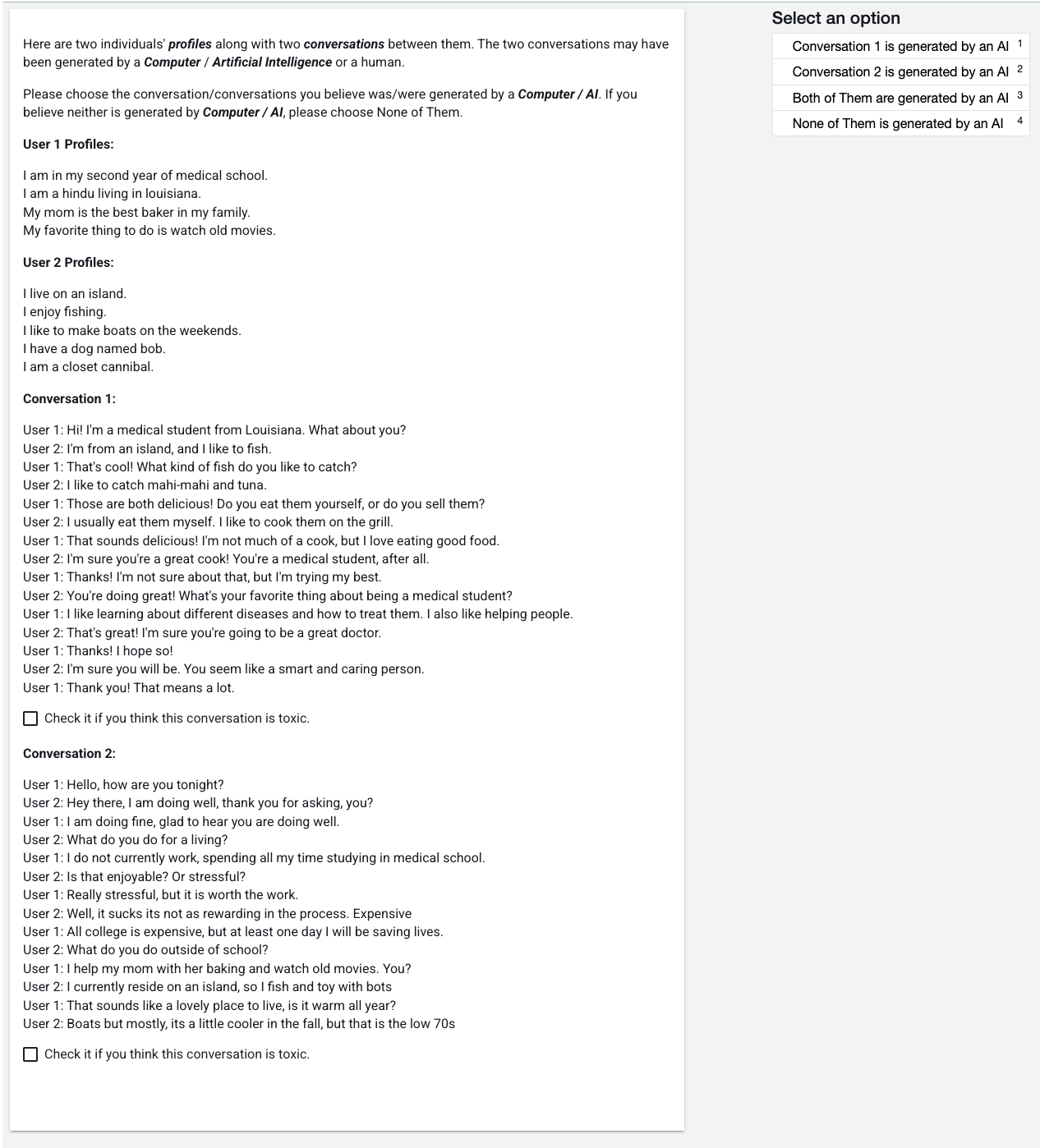}
    \caption{\small Preview of the Turing Test Task on the Crowdsourcing Platform}
    \label{fig:h_eval}
\end{figure*}

\paragraph{Turing Test}
We randomly select 200 user pairs from \personachatabbr{}.
For each example, we show the annotators the user pair, together with the corresponding conversations from \personachatabbr{} and \ourdatasetabbr{}, and ask them to select the conversation that was synthetically generated.
We show an example of this crowdsourcing task in Figure \ref{fig:h_eval}.
The results of the Turing test are available in Table \ref{tab:annotator_agreement}.
We report the losing rate of \ourdatasetabbr{} in Turing test, and Fleiss' Kappa to assess the inter-rater agreement.
The agreement falls into the fair to moderate agreement bucket. 

\begin{figure*}
    \centering
    \includegraphics[scale=0.5]{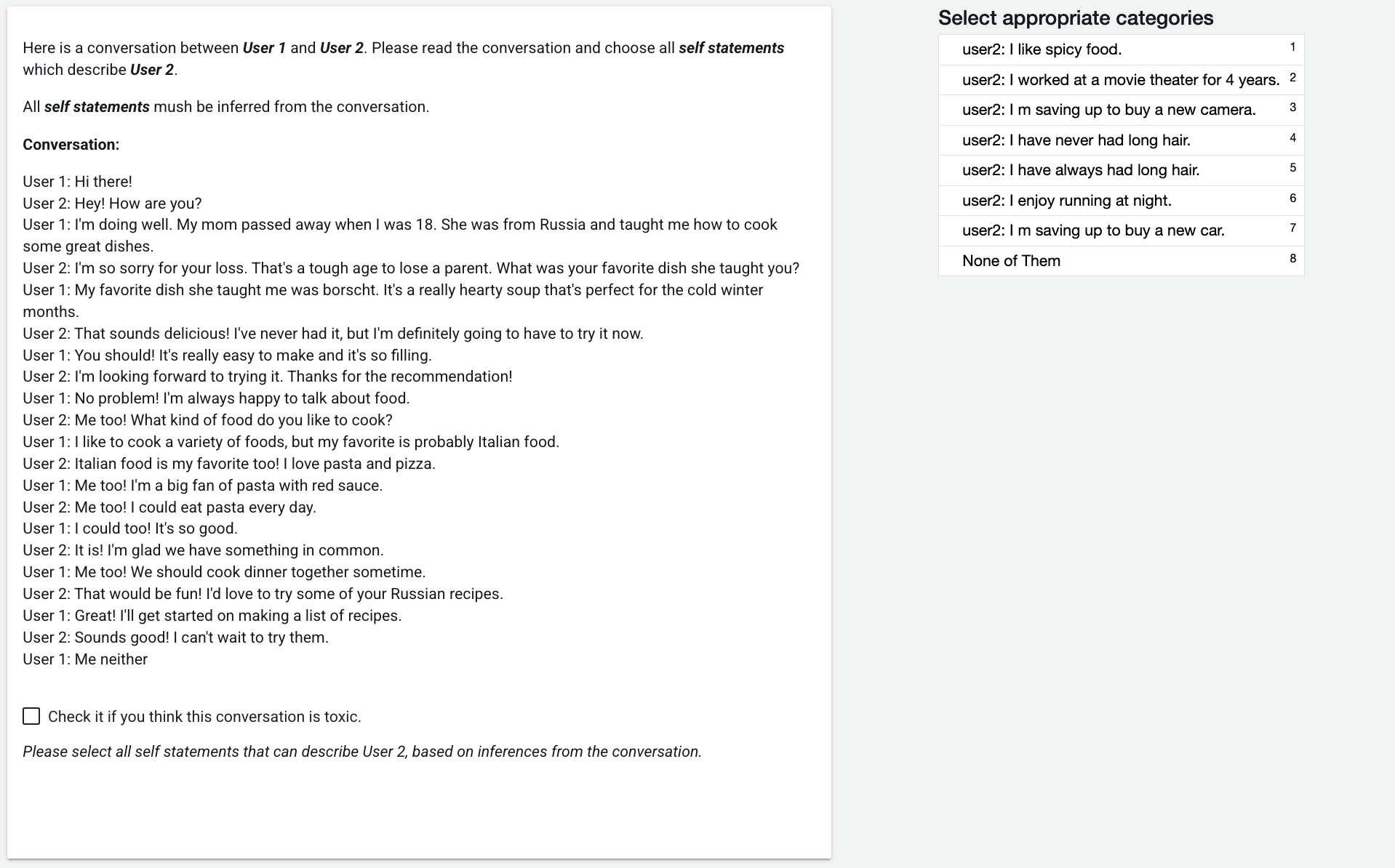}
    \caption[width=\textwidth]{Preview of the Faithfulness Task on the Crowdsourcing Platform.}
    \label{fig:h_fact}
\end{figure*}

\begin{table}
    \centering
    \begin{small}
    \begin{tabular}{c|c|c|c}
        \toprule
        Conversation Source &  $\%$ Lose & $\kappa$ & $\#$ annotators\\  \midrule
        \ourdatasetabbr{} Iter 1 & 17.2 & 0.41 & 50\\ %.364  \\
        \ourdatasetabbr{} Iter 2 &  18.5 &  0.48 & 40\\ %.427  \\
        \ourdatasetabbr{} Iter 3 &  8.8 & 0.22 & 11\\%.428  \\
        \hline
        \ourdatasetabbr{} Iter 3* & 8.04  & 0.56 & 24\\
        \ourdatasetabbr{} (LLM2) & 11.5 & 0.49 & 36\\ 
        \bottomrule
    \end{tabular}
    \caption{\small
    Turing test results on a sample of 200 conversations. The first column shows the percentage of \ourdatasetabbr{} losing compared to \personachatabbr{} in the Turing test. Note that the last iteration (3) of \ourdatasetabbr{} is an evaluation of the segment of conversations based on the extended persona set.
    }
    \label{tab:annotator_agreement}
    \end{small}
\end{table}
\paragraph{Faithfulness}
We present the annotators with a conversation, and a set of options of persona attributes.
The annotators are asked to select the user persona attributes they would infer from the conversation. 
Figure \ref{fig:h_fact} shows a sample of the annotation task in this study.
The options include the persona attributes of the speakers in the conversation, and a set of distractor persona attributes.
We created distractor persona attributes using different strategies to cover different difficulty levels.
For a persona attribute set $\Pi$, we create a set $\neg \Pi$ of distractor persona attributes as:

\textbf{Negated personas} We prompt an LLM to negate persona attributes. For example, the negation of persona attribute "I like vegetables" is "I don't like vegetables". 

\textbf{Random personas} We randomly select persona attributes from user profiles in other conversations in the dataset. 

\textbf{Contradicting personas} We prompt an LLM to generate a persona attribute which contradicts the users' personas.

Each entry of this task includes 8 user persona attributes as options, where 4 of them are the real persona attributes, and the other 4 are distractors.
We evaluate the precision of the human annotators, and report it as a proxy to the conversation faithfulness in Table \ref{tab:human_eval_iter}. 

% \subsection{Dialog Quality Evaluation}

\section{Ablation Studies}
We run several ablation studies to evaluate the importance of individual components in our framework.
We begin by analyzing the effect of the persona expansion module.
We then review the impact of each expert in the mixture forming our Critic.

\subsection{Persona Expansion}
We assess the importance of the query-based persona expansion module introduced in Section \ref{sec:pe}.
Similarly to the experiment outlined in Section \ref{sec:peq}, we run the persona expansion on two datasets: Wikipedia and \personachatabbr{}.
The results of this experiment are presented in Table \ref{tab:ablation_persona_expansion}.
We designate the persona expansions without the inducted query set (Q) as `Wikipedia-0', and `\personachatabbr{}-0', and run the same number of iterations for each (100 iterations).
We observe that \personachatabbr-0 includes 4,477 new persona attributes, 20 percent less than \personachatabbr{}.
The difference in the number of newly generated persona attributes is more pronounced in the case of Wikipedia, where Wikipedia-0 consists of 4,742 persona attributes, 50 percent less than Wikipedia+.
This trend is also observed in the number of persona clusters, with \personachatabbr{}-0 and Wikipedia-0 having $6\%$ and $49\%$ less clusters respectively. 
This pattern suggests the effectiveness of the query-based persona expansion in maintaining the diversity of the persona set.
Furthermore, the average persona attribute length in \personachatabbr{}-0 is 11.38 tokens, which is $28\%$ less than \ourdatasetabbr{}.
This reduction points to less detailed and specific persona attributes.
In contrast, the expansion in `Wikipedia-0' exhibits similar average persona attribute lengths compared to `Wikipedia+'.
\begin{table*}
    \centering
    \begin{small} 
    \resizebox{\textwidth}{!}{
    \begin{tabular}{c|ccc|ccc}
    \toprule
    Dataset  & \personachatabbr & \ourdatasetabbr & \personachatabbr-0 & Wikipedia & Wikipedia+ & Wikipedia-0 \\ 
    \midrule

    $\#$ Persona Attributes & 4,723 & \textbf{10,371} & 9,200 & 8,768 & \textbf{18,293} & 13,510\\
    $\#$ Clusters & 323 & \textbf{553} & 520  & 408 & \textbf{986} & 502\\
    InterCluster-Dist & 0.836 & 0.863& 0.842  &0.816 & 0.85 & 0.83\\
    % Distance & & & & & &\\ 
    AVG length & 7.65 & $\textbf{15.9}^{*}$ & $11.38^*$& 10.45 & $\textbf{15.2}^{*}$ & $\textbf{15.2}^*$\\

    \bottomrule
    \end{tabular}
    }
    \caption{Evaluation of the Expanded Persona Attribute Sets. The numbers with $'*'$ indicate the metric value on the newly generated persona attributes, in contrast to the initial persona attributes.}
    
    \label{tab:ablation_persona_expansion}
    \end{small}
    \vspace{-5mm}
\end{table*}
\subsection{Conversation Quality}
\label{convq}
We analyze the effect of the experts within our Critic.
We remove each expert, and generate a dataset using one iteration of our framework.
We compare the resulting datasets against the output of the first iteration of \ourdatasetabbr{}.
We use the evaluators introduced in \ref{llmeval-sec}.
The results of this experiment are summarized in Table \ref{tab:auto-eval}.
We observe that the exclusion of the experts results in worse performance according to most criteria: 3 out of 4 in LLM-Eval, 4 out of 6 in GPT-Score, and 3 out of 5 in G-Eval.
% Using LLM-Eval as the evaluator,  Context, Grammar and Relevance,  \ourdataset{}-Iter 1 get higher quality score. 
% Using GPT-Score as the evaluator,  Context, Grammar and Relevance,  \ourdataset{}-Iter 1 get higher quality score. 

\subsection{Faithfulness}
We ablate the faithfulness critic, and generate a dataset that we compare against \ourdatasetabbr{}.
We compare these datasets both automatically, using human annotators (Turing Test), and using a prompted LLM (LLM-Evaluator).
We describe this study in more details below.

\paragraph{Turing Test}
We run a human study to compare a small subset of conversations created without the faithfulness expert against their equivalent created with that expert.
This experiment process is similar to \ref{exp:h_faith} and it is conducted for 200 conversations.
The precision decreases from $78.0\%$ to $66.0\%$ without this critic, highlighting its effectiveness in eliminating conversations with contradictory information about user personas.
The recall decreases from $36.0\%$ to $23.0\%$, demonstrating a higher reflection of personas in the conversations in the presence of the faithfulness expert.

\paragraph{LLM-Evaluator}
We extend our comparison to the entire dataset using an LLM as an annotator, following \cite{He2023AnnoLLMML, Bansal2023LargeLM, Chiang2023CanLL}.
Table \ref{tab:abl_faith} shows the faithfulness of the conversations generated in the first iteration without the faithfulness expert.
The templates used in the LLM-based annotators are described in Table \ref{tab:auto-eval} in the rows with "LLM-Faithfulness" as their evaluator.
Note that the annotator-based LLM is created using a different LLM, gpt-3.5-turbo \cite{Brown2020LanguageMA, Ouyang2022TrainingLM}, than the LLM used for dataset generation. 
\begin{table*}
    \centering
    \begin{tabular}{c|cc|cc}
    \toprule
        & \multicolumn{2}{c|}{LLM Evaluator (\%)} & \multicolumn{2}{c}{Human Evaluator (\%)} \\
        Absent Component & Inference & Contradiction  & Precision & Recall  \\ \midrule
        % None & 34.05 &28.05 & 78.0 & 36.0 \\
        % Faithfulness& 33.6 & 28.9 &  66.0 & 23.0 \\
        % FED & 32.2 & 28.55  & & \\
        None & \textbf{33.2} & \textbf{24.5} & \textbf{78.5} & \textbf{36.4} \\
        Faithfulness & 32.7 & 28.8 & 66.1 & 23.1     \\
        FED & 31.7 & 28.5 & N/A & N/A\\
        \bottomrule
    \end{tabular}
    \caption{\small 
    Faithfulness of Generated Conversation Datasets Using the Framework While Eliminating Each Component. The first row represents the framework without removing any component, equivalent to the first iteration of \ourdataset{}.}
    \label{tab:abl_faith}
\end{table*}

% Edit is done till this point
\subsection{Next Utterance Prediction}
We follow the experimental setting described in section \ref{sec:nup}, and compare the performance of various next utterance prediction models trained on \ourdatasetabbr{} against the same models trained on datasets created in the absence of certain experts.

When using the IR Baseline as the next utterance prediction method, we observee that its highest performance of 39\% hit@1 occurs when the FED critic is absent during dataset creation.
This outcome aligns with FED's emphasis on conversation quality, excluding persona-related aspects.
Conversely, the Transformer Ranker, capable of understanding intricate concepts, achieves its peak performance of 13.9\% hit@1 when none of the experts are absent.
This result supports the inclusion of both FED and the Faithfulness expert in the model architecture.
In generative models, the absence of FED impacts the next utterance prediction model the most, leading to a notable decline in performance (e.g. $-12\%$ hit@1, $-9\%$ BLEU, $-10\%$ ROUGE).
This observation underscores the crucial role played by FED in enhancing the generative capabilities of the model.

\begin{table*}
    \centering
    
    \resizebox{\textwidth}{!}{
    \begin{tabular}{c|c|ccc|ccc|ccc}
    \toprule
        \multicolumn{2}{c|}{Absent Component} & \multicolumn{3}{c|}{Faithfulness}& \multicolumn{3}{c|}{FED} & \multicolumn{3}{c}{None}\\ \midrule
        
        Method & Metric & None   &  Persona & $\%$ Change & None   &  Persona & $\%$ Change & None  & Persona &  $\%$ Change \\ \midrule
        IR Baseline & hit@1 & 18.7 & 38.7 & \textbf{+106} &
        
        19.0 & \textbf{39.0} & +105 &
        18.9  & 38.7 &  +105\\
        % & hit@5 & 43.5 &	60.7 &	+39.4 & 43.7&	\textbf{60.8} &	+39.0 & 43.7	&60.7&	+38.8 \\
       Transformer (Ranker)
         & hit@1 & 10.9 & 13.5 & +24 &
         10.7 & 13.6 & \textbf{+27} &
         12.4   & \textbf{13.9}  & +11 \\
         
        %  (Ranker) & hit@5 & 39.0 &	43 &	+10.1 & 43.2 &	\textbf{46.5} &	+7.6 & 40.4 & 45.4 & +12.4 \\
        \midrule

         & hit@1 & \textbf{8.9} & 7.4 & -16 &
          8.4 & 7.4 & \textbf{-12} &
         8.2 & 7.0 & -14 \\
        %  & hit@5 &
        %  32.2 &	\textbf{33.1} &	+2.7&
        %  33.0&	32.8 &	-0.5 &
        % 32.3	& 32.8&	+1.5 
        %  \\
         
         Transformer&  Perplexity & 204 & 214 & +5 &
         \textbf{174} & 185  & +6 &  
         203 & 210 & \textbf{+3}\\
        
        (Generator) & BLUE & 0.11 &  0.10 & -11 &
        0.11 & 0.10 & \textbf{-9} & 
        0.10 & 0.08 & -15\\
        
        & ROUGE & 0.14 & 0.15 & -12&
        0.14 &  0.12 & \textbf{-10} & 
         0.13 & 0.10 & -17 \\

        \bottomrule
    \end{tabular}
    }
    % \end{small}
    \caption{\small 
    Results of the Next Utterance Prediction Experiment in the Ablation Study. The numbers in the table represent the performance of the trained model on the test portion of the \personachat{} dataset.
    }
    \label{tab:abl-nup}
    % \end{small}
    \vspace{-5mm}
\end{table*}

\end{document}